%% file: eccv2022submission.tex
\documentclass[runningheads]{llncs}
\usepackage{graphicx}

\usepackage{tikz}
\usepackage{comment}
\usepackage{amsmath,amssymb} %
\usepackage{color}
\usepackage{dcolumn}
\usepackage{cite}
\usepackage{url}

\usepackage[accsupp]{axessibility}  %

\usepackage{enumitem}
\usepackage{caption}

\def\eg{\emph{e.g.}}

\def\etal{\emph{et al.}}

\usepackage{color}

\begin{document}
\pagestyle{headings}
\mainmatter

\title{NeRF for Outdoor Scene Relighting}

\titlerunning{NeRF-OSR}
\author{Viktor Rudnev$^{1,2}$ \and
Mohamed Elgharib$^{1}$ \and
William Smith$^{3}$ \and
Lingjie Liu$^{1}$ \and
Vladislav Golyanik$^{1}$ \and
Christian Theobalt$^{1}$
}

\authorrunning{V. Rudnev et al.}
\institute{$^1$MPI for Informatics, SIC $\quad$ $^2$Saarland University, SIC $\quad$ $^3$University of York} 

\maketitle

\input{figures/teaser}
\input{0_abstract}
\input{1_intro}

\input{2_related}
\input{3_method_ye}
\input{4_dataset}

\input{5_results_ye}

\input{6_discussion}

\input{7_conclusion}

\bibliographystyle{splncs04}
\bibliography{eccv2022submission}

\setcounter{section}{0}
\renewcommand\thesection{\Alph{section}}
\newcommand{\suppsection}{\subsection}
\clearpage

\title{NeRF for Outdoor Scene Relighting}
\subtitle{--Supplementary Material--} 

\titlerunning{NeRF-OSR}
\author{Viktor Rudnev$^{1,2}$ \and
Mohamed Elgharib$^{1}$ \and
William Smith$^{3}$ \and
Lingjie Liu$^{1}$ \and
Vladislav Golyanik$^{1}$ \and
Christian Theobalt$^{1}$
}

\authorrunning{V. Rudnev et al.}
\institute{$^1$MPI for Informatics, SIC $\quad$ $^2$Saarland University, SIC $\quad$ $^3$University of York} 

\makeatletter

\newpage
 \markboth{}{}%
 \def\lastand{\ifnum\value{@inst}=2\relax
                 \unskip{} \andname\
              \else
                 \unskip \lastandname\
              \fi}%
 \def\and{\stepcounter{@auth}\relax
          \ifnum\value{@auth}=\value{@inst}%
             \lastand
          \else
             \unskip,
          \fi}%
 \begin{center}%
 \let\newline\\
 {\Large \bfseries\boldmath
  \pretolerance=10000
  \@title \par}\vskip .8cm
\if!\@subtitle!\else {\large \bfseries\boldmath
  \vskip -.65cm
  \pretolerance=10000
  \@subtitle \par}\vskip .8cm\fi
 \setbox0=\vbox{\setcounter{@auth}{1}\def\and{\stepcounter{@auth}}%
 \def\thanks{}\@author}%
 \global\value{@inst}=\value{@auth}%
 \global\value{auco}=\value{@auth}%
 \setcounter{@auth}{1}%
{\lineskip .5em
\noindent\ignorespaces
\@author\vskip.35cm}
 {\small\institutename}
 \end{center}%
\makeatother

\vspace{2em}
\input{supplemental}

\end{document}

%% file: figures/teaser.tex
\begin{center}
    \centering
    \begin{tabular}{@{}c@{~}c@{~}c@{~}c@{}}
       \includegraphics[height=2.05cm,clip=true]{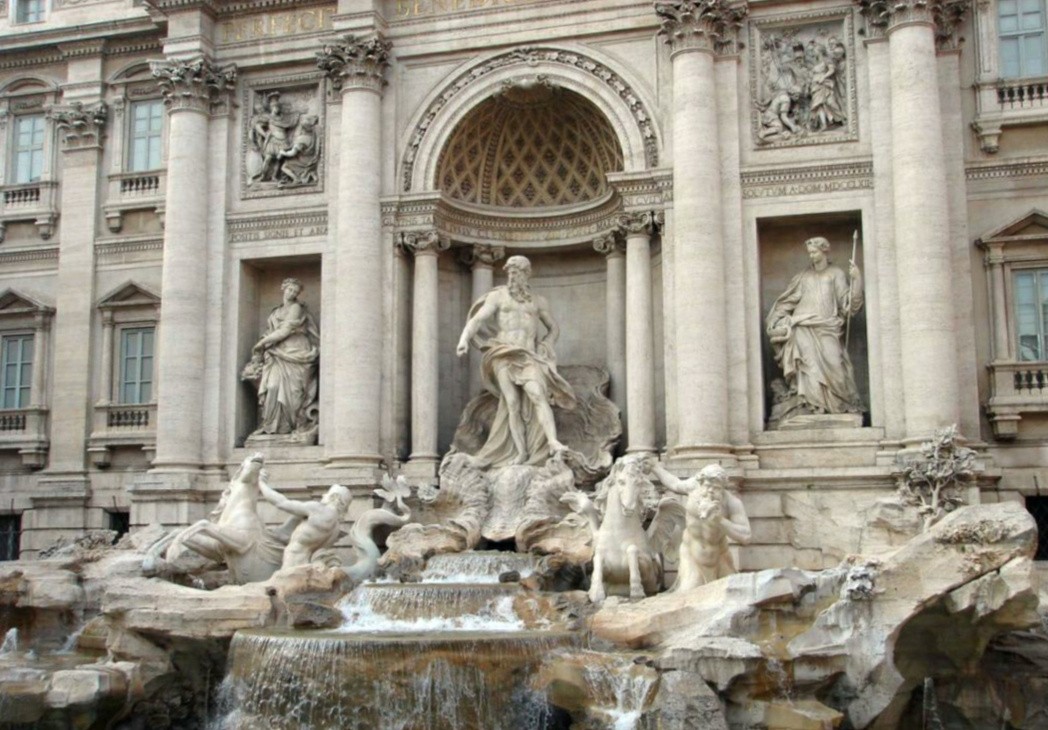}  &  
       \includegraphics[height=2.05cm,clip=true]{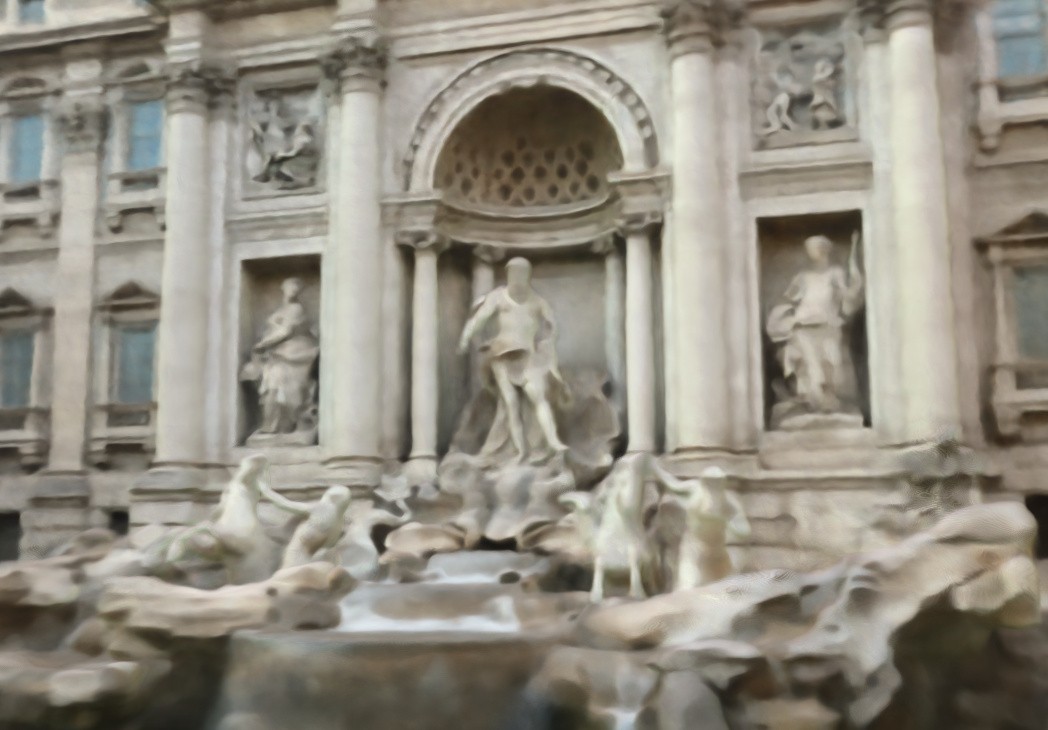} &
       \includegraphics[height=2.05cm,clip=true]{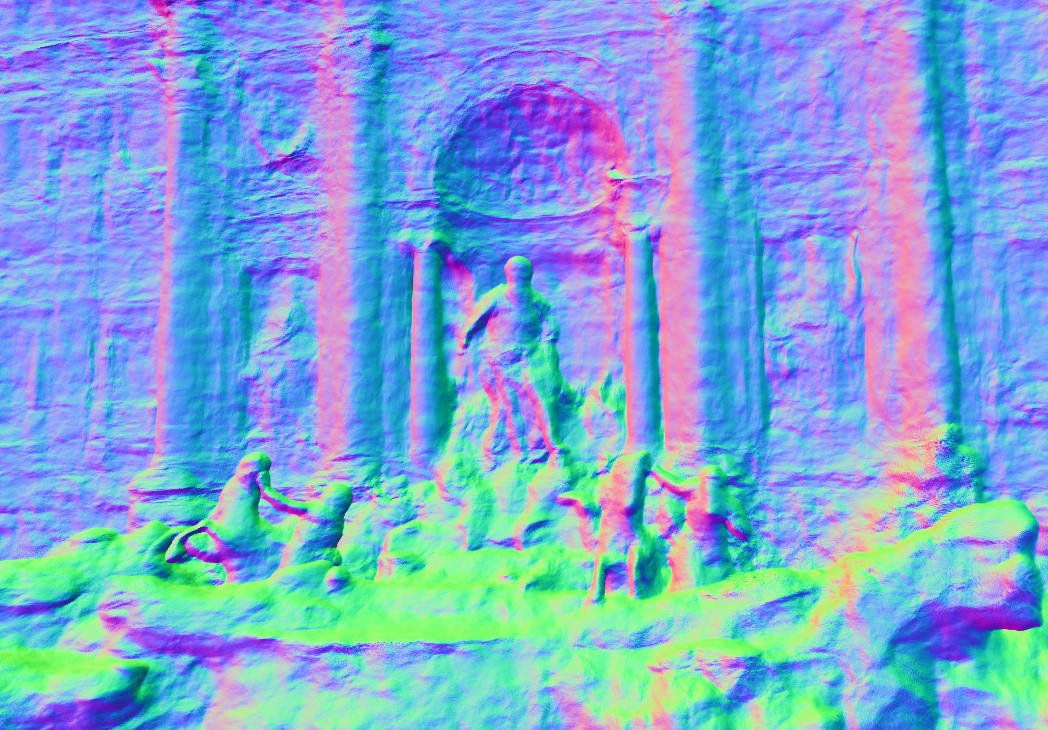} &
       \includegraphics[height=2.05cm,clip=true]{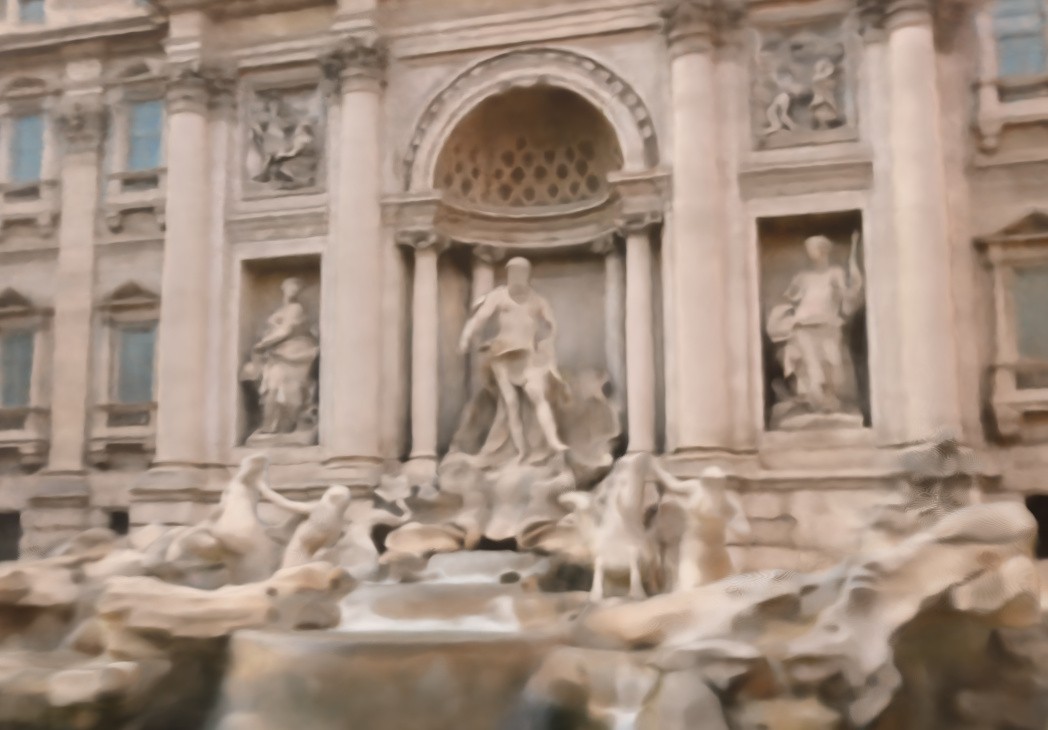} \\
       \small{Training image}  & \small{Reconstruction} & \small{Surface normals} & \small{Diffuse Albedo} \\[0.4em]
       \includegraphics[height=2.05cm,clip=true]{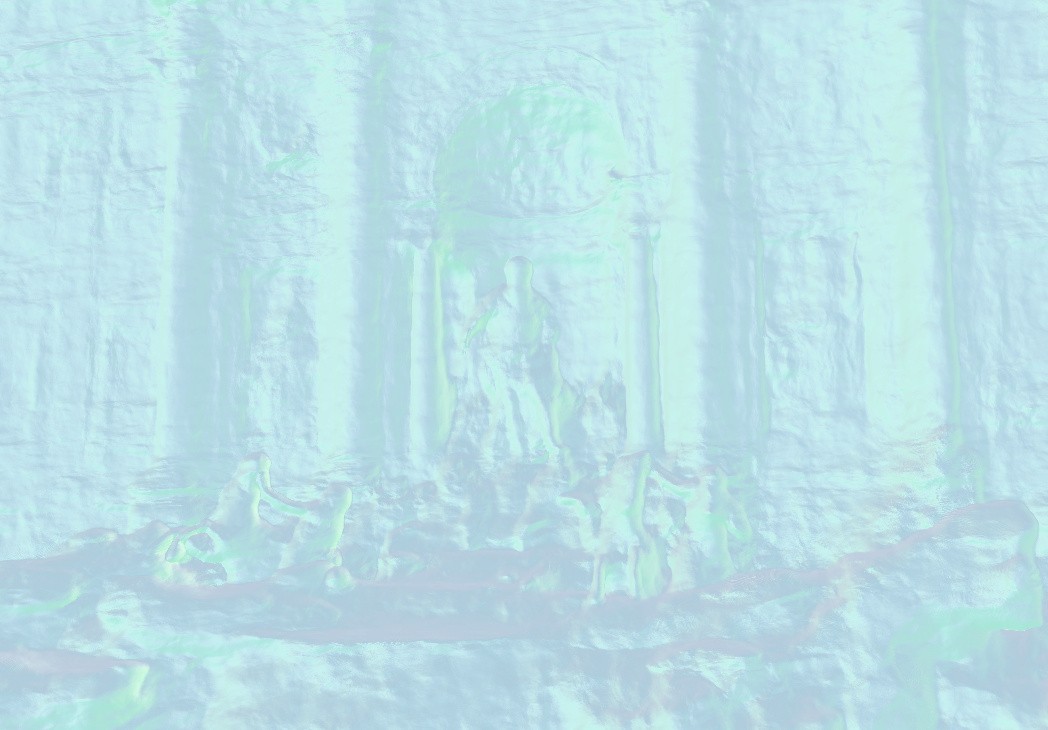}       &
       \includegraphics[height=2.05cm,clip=true]{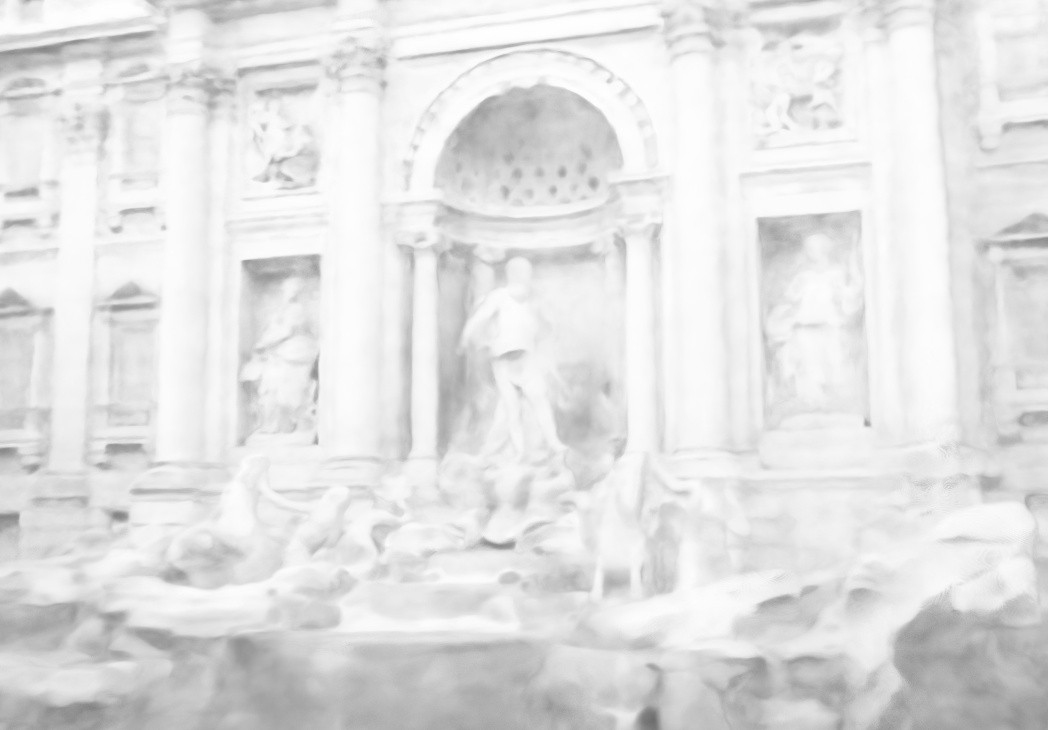} &
       \includegraphics[height=2.05cm,clip=true]{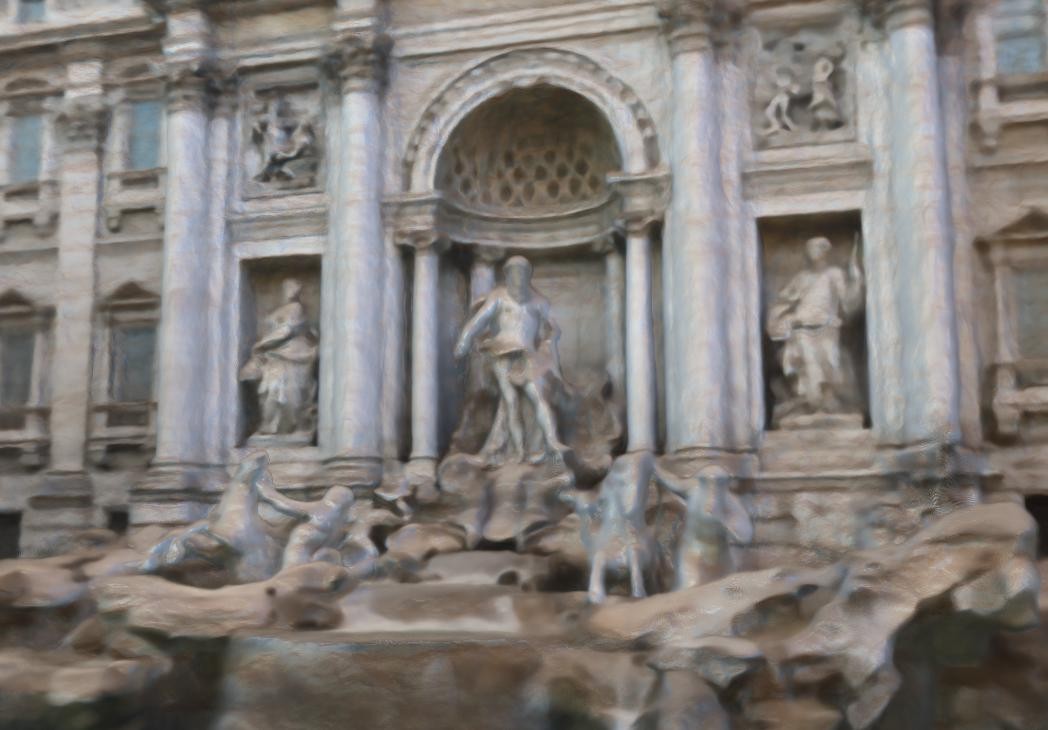} &
      \includegraphics[height=2.05cm,clip=true,trim=34px 0px 34px 0px]{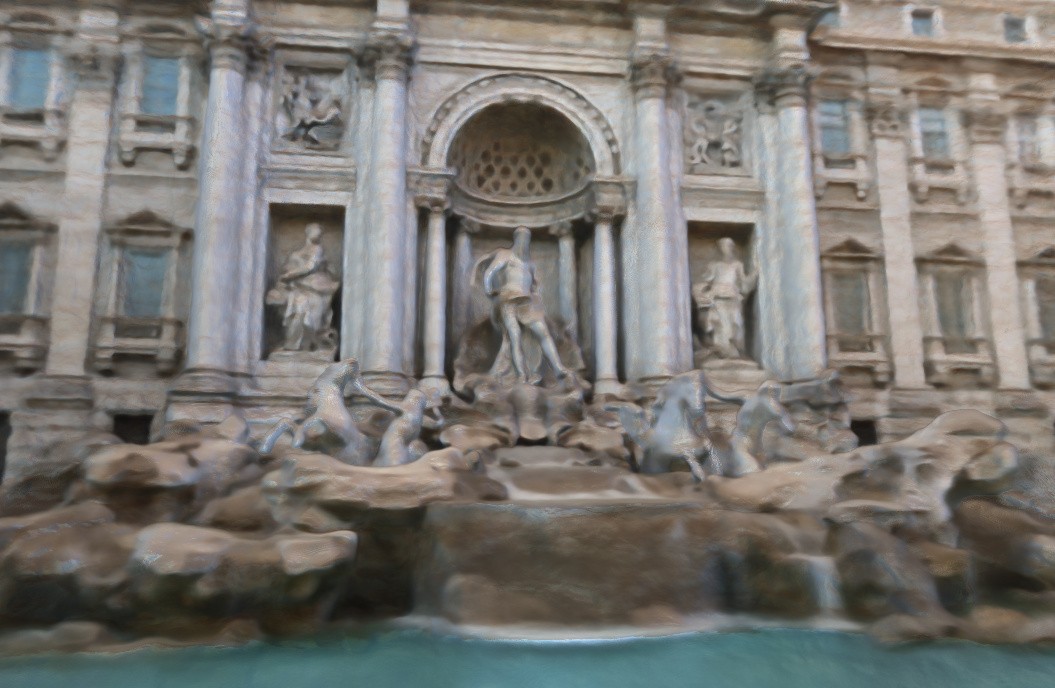}\\
       \small{Illumination} & \small{Shadows} & \small{Novel lighting (N. l.)} & \small{N. l. and viewpoint} \\
    \end{tabular}
    
    \captionof{figure}{\textbf{NeRF-OSR is the first neural radiance fields approach for outdoor scene relighting.} 
    We learn a neural representation of the scene geometry, diffuse albedo and illumination-dependent shadows from a set of images capturing the same site from different viewpoints and at different times.
    The learnt intrinsics enable simultaneous editing of both the scene's lighting and viewpoint.%
    } 
    \label{fig:teaser}
\end{center}%

%% file: 0_abstract.tex
\begin{abstract}
Photorealistic editing of outdoor scenes from photographs requires a profound understanding of the image formation process and an accurate estimation of the scene geometry, reflectance and illumination. 
A delicate manipulation of the lighting can then be performed while keeping the scene albedo and geometry unaltered. We present \mbox{NeRF-OSR}, \textit{i.e.,} the first approach for outdoor scene relighting based on neural radiance fields. In contrast to the prior art, our method allows simultaneous editing of illumination and camera viewpoint using only a collection of outdoor photos shot in uncontrolled settings. Moreover, it enables direct control over the scene illumination, as defined through a spherical harmonics model. 
For evaluation, we collect a new benchmark dataset of several outdoor sites %
photographed from multiple viewpoints and at different times\footnote{see the project web page \url{https://4dqv.mpi-inf.mpg.de/NeRF-OSR/}}. 
For each time, a $360^\circ$ environment map is captured together with a colour-calibration chequerboard to allow accurate numerical evaluations on real data against ground truth. 
Comparisons against SoTA show that \mbox{NeRF-OSR} enables controllable lighting and viewpoint editing at higher quality and with realistic self-shadowing reproduction. 
\end{abstract}

%% file: 1_intro.tex
\section{Introduction}

Controllable lighting editing of real scenes from photographs is a long-standing and challenging problem, with several applications in virtual and augmented reality~\cite{Meka20,Guo19,Yu20,Philip19,Meka2018}.
It requires explicit modelling of the image formation process and an accurate estimation of the material properties and scene illumination. 
Such scene decomposition enables manipulating the lighting in isolation while maintaining the integrity of the remaining scene components (\textit{e.g.,}~albedo  and geometry.)
While several methods for controllable lighting editing exist, some solutions are dedicated to a specific class of objects such as human faces~\cite{mallikarjun2021photoapp,Sun19} and human bodies~\cite{Meka20,Guo19}. 
Other solutions are designed for processing either indoor~\cite{zhang2021nerfactor,Srinivasan21NeRV,Xu18deep,Sengupta19ICCV,Garon19,Meka2018} or outdoor~\cite{Xing13,Duchene15,Barron15,Philip19,Yu20} scenes. 
Due to the very different nature of indoor and outdoor data, methods for relighting them were largely treated separately in the literature. 
In this work, we focus on outdoor scene relighting. Unlike existing methods~\cite{Xing13,Duchene15,Barron15,Philip19,Yu20}, our approach is the first to simultaneously edit both scene illumination and camera viewpoint.

The recently proposed Neural Radiance Fields  (NeRF)~\cite{Mildenhall20nerf} is a powerful neural 3D scene representation capable of self-supervised training from 2D images recorded by a calibrated monocular camera \cite{KaiZhang2020,park2020nerfies,tretschk2021nonrigid,2021arXiv211105849T}. 
At test time, NeRF can produce photorealistic novel scene views. 
While there were a few attempts to extend NeRF for lighting editing~\cite{Brualla20nerfw,Srinivasan21NeRV,Boss20NeRD,sun2021nelf,zhang2021nerfactor}, existing approaches are either designed for a specific object class~\cite{sun2021nelf}, require known or single illumination condition for training~\cite{Srinivasan21NeRV,zhang2021nerfactor} or they do not model important outdoor illumination effects such as cast shadows~\cite{Boss20NeRD}. 
Most existing NeRF-based relighting methods~\cite{Srinivasan21NeRV,Boss20NeRD,sun2021nelf,zhang2021nerfactor} are not designed for outdoor scenes captured in uncontrolled settings.
An exception to this is, at first sight, NeRF in the Wild (NeRF-W) \cite{Brualla20nerfw} trained from uncontrolled images, factoring per-image appearance into an embedding space. 
However, NeRF-W and more recent  follow-ups~\cite{chen2022hallucinated,tancik2022blocknerf} do not perform intrinsic image decomposition and thus semantically meaningful parametric control of lighting, shadows or even albedo is not possible. 

This paper addresses the shortcomings of existing methods and presents NeRF-OSR, \textit{i.e.,} the first approach based on neural radiance fields that can change both illumination and camera viewpoint of outdoor scenes photographed in uncontrolled settings, in a high-quality and semantically meaningful way; see Fig.~\ref{fig:teaser}. 
Our approach models the image formation process, disentangling the input image into its intrinsic components and scene illumination. 
It also contains a dedicated network for learning shadows, whose realistic reproduction is crucial for high-quality outdoor scene relighting. NeRF-OSR is trained in a self-supervised manner on multiple images of a site photographed from different viewpoints and under different illuminations.  
We evaluate our method qualitatively and quantitatively on a variety of outdoor scenes and show that it outperforms state of the art. 
Aspects of the novelty of our work include: 
\begin{itemize}[leftmargin=*]
    \item NeRF-OSR, \textit{i.e.}, the first method using neural radiance fields for outdoor scene relighting supporting simultaneous and semantically meaningful editing of scene illumination and camera viewpoint. Our model has explicit control over the scene intrinsics, including local shading, shadows and even albedo.
    \item Our method learns a neural scene representation that  decomposes the scene into spatial occupancy, illumination, shadowing and diffuse albedo reflectance. It is trained in a self-supervised manner from outdoor data captured from various viewpoints and at different illuminations.  
    \item A new and biggest in literature benchmark dataset for outdoor scene relighting. 
    It includes eight buildings photographed from 3240 viewpoints and at 110 different times. In addition, it is the first one that includes  colour-calibrated $360^\circ$ environment maps, which allows accurate numerical evaluations.
\end{itemize}

%% file: 2_related.tex
\section{Related Work} 
\label{sec:related}

\textbf{Scene Relighting.} 
There are several methods for outdoor illumination editing~\cite{Sunkavalli07,Lalonde09,Karsch11,Laffont12,Xing13,Duchene15,Barron15,Yu19,Yu20,Philip19}.
Some of them focus on integrating objects into images in an illumination-consistent manner~\cite{Karsch11,Xing13}, while others process the full scene  ~\cite{Sunkavalli07,Lalonde09,Laffont12,Duchene15,Barron15,Philip19,Yu19,Yu20}. 
Duchene~\etal~\cite{Duchene15} estimate scene reflectance, shading and visibility from multiple views shot at fixed lighting. 
They produce novel relighting effects such as moving cast shadows. 
Barron~\etal~\cite{Barron15} formulate inverse rendering through statistical inference. 
Given a single RGB image of an object, their method
estimates the most likely 
shape normals, reflectance, shading and illumination that can reproduce the examined image. 
They assume piecewise smooth and low-entropy reflectance images and isotropic (with frequent bends) surfaces. 
Philip~\etal~\cite{Philip19} guide relighting via a proxy geometry estimated from multi-view images. 
Their neural network translates image-space buffers of the examined scene into the desired relighting.  
The buffers include shadow masks (estimated from the extracted geometry), normal maps and illumination components. 
Philip~\etal's method is trained with high-quality synthetic data. 
However, it is primarily designed to edit only the illumination of the input and not the camera viewpoint. 
Furthermore, their illumination model is limited to sun lighting and can not handle other cases such as cloudy skies. 

Yu and Smith~\cite{Yu19} 
estimate the albedo, normals and lighting of an outdoor scene from just a single image; lighting is modelled through spherical harmonics (SH) with a statistical model as a prior. 
Relighting is then achieved by editing the reconstructed illumination (using a low-frequency model). 
Yu~\etal~\cite{Yu20} train a method for scene relighting given a single image in a self-supervised manner on a large corpus of uncontrolled outdoor images. 
A neural renderer takes the original albedo and geometry, the target shading and the target shadowing, and relights the scene; 
a dedicated network predicts the target shadows. 
Next, residuals of the inverse rendering are also supplied to the neural renderer as input to better capture scene details. 
Impressive results are shown visually and validated numerically on a new benchmark dataset. 
In contrast to our approach, neither Yu~\etal~\cite{Yu20} nor Yu and Smith~\cite{Yu19} can edit the camera viewpoint.

Recently, there were efforts in developing relighting methods using NeRF  backbone~\cite{Srinivasan21NeRV,Boss20NeRD,sun2021nelf,zhang2021nerfactor}. 
Most of these methods operate in a setting different from ours, \textit{i.e.,} they either require input images with a single illumination condition~\cite{zhang2021nerfactor}, assume a known illumination during training~\cite{Srinivasan21NeRV} or are designed for a specific class of objects such as  faces~\cite{sun2021nelf}. 
The closest to our technique is NeRD by Boss~\etal~\cite{Boss20NeRD}, in the sense it can operate on images of the same scene shot under different illuminations. 
Here, the spatially varying BRDF of the examined scene is estimated through the help of physically-based rendering. 
To allow fast rendering at arbitrary viewpoints and illumination, the learnt reflectance volume is converted into a relightable texture mesh. 
Unlike our NeRF-OSR, NeRD does not explicitly model shadows, which are crucial for high-quality outdoor scene relighting. 
Furthermore, it requires the examined object to be at a similar distance from all views---an assumption that can not be easily satisfied for outdoor photographs captured in an uncontrolled setup.

\textbf{Style-based Editing.} Scene relighting techniques are distantly related to 
style-based category of appearance editing methods~\cite{Shih13,Luan17,Meshry19,Li20crowdsampling,Nam19,Brualla20nerfw,chen2022hallucinated,tancik2022blocknerf}. 
Unlike relighting methods, the latter do not have a physical understanding of the scene illumination and
seek to edit the overall appearance at once.
Hence, they lack explicit parametric control over the local shading and shadows. 
In contrast, our NeRF-OSR performs scene intrinsic decomposition and seeks to edit illumination in isolation from albedo and geometry. 
It also directly models illumination-based shadows, which is crucial for high-quality outdoor relighting. 
Next, our intrisic decomposition allow editing applications that are not possible by style-based methods by any means (\textit{e.g.,} inserting objects by editing the albedo channel separately and then relighting the entire composited new scene). 

%% file: 3_method_ye.tex
\section{Method}

NeRF-OSR takes as input multiple RGB images of a single scene, shot at different timings and from different viewpoints. 
It then renders the examined scene from an arbitrary viewpoint and under various illuminations. 
Our method estimates the scene intrinsics explicitly and has direct access to the scene illumination.
It also includes a dedicated component for predicting shadows, \textit{i.e.,} an essential feature of outdoor scene illumination. 

\input{figures/pipeline}

An overview of NeRF-OSR is shown in  Fig.~\ref{fig:pipeline}. 
At its heart is a neural radiance fields (NeRF), \textit{i.e.,} a neural implicit scene representation for volumetric rendering. 
Our method is trained in a self-supervised manner on outdoor data captured in uncontrolled settings and can render photorealistic views. 
Next, we describe in Sec.~\ref{ssec:original_NeRF} the NeRF model~\cite{Mildenhall20nerf} without view-dependent effects, which we build upon. 
We then discuss our illumination model and how it is adapted in a volumetric-based representation in  Secs.~\ref{ssec:NeRF_sph_harmonics}--\ref{ssec:shadow_network}. 
The objective function is presented in Sec.~\ref{ssec:objective_function}, followed by a discussion of the training details (Sec.~\ref{ssec:training_details}). 

\subsection{Neural Radiance Fields  (NeRF)}\label{ssec:original_NeRF} 

For each point $\mathbf{x}$ in 3D space,  NeRF~\cite{Mildenhall20nerf} defines its density  $\sigma(\mathbf{x})$ and colour $\mathbf{c}(\mathbf{x})$. 
To render an image, a ray is cast from the camera origin $\mathbf{o}$, in a direction $\mathbf{d}$ corresponding to each of the output pixels. 
$N_{depth}$ points $\{\mathbf{x}_i\}_{i=1}^{N_\mathrm{depth}}$ are sampled along each ray, where $\mathbf{x}_i=\mathbf{o}+t_i\mathbf{d}$ and $\{t_i\}_{i=1}^{N_\mathrm{depth}}$ are the corresponding ray depths.
The final colour in the image space  $\mathbf{C}(\mathbf{o}, \mathbf{d})$ 
is obtained by integrating the density and colour along the ray $(\mathbf{o}, \mathbf{d})$ as follows: 
\begin{equation}
\small 
    \label{eq:rendereq}
    \mathbf{C}(\mathbf{o}, \mathbf{d})=
    \mathbf{C}\left(\{\mathbf{x}_i\}_{i=1}^{N_\mathrm{depth}}\right)=
    \sum_{i=1}^{N_\mathrm{depth}}T(t_i)\alpha(\sigma(\mathbf{x}_i)\delta_i)\mathbf{c}(\mathbf{x}_i),
\end{equation}
where $T(t_i)=\exp\left(-\sum_{j=1}^{N_\mathrm{depth}-1}{\sigma(\mathbf{x}_j)\delta_j}\right)$, $\delta_i=t_{i+1}-t_i$, and $\alpha(y)=1-\exp(-y)$.
The depths $\{t_i\}_{i=1}^{N_\mathrm{depth}}$ are selected using stratified sampling from the uniform distribution, spanning
the depths along $(\mathbf{o}, \mathbf{d})$ starting from the near and ending at the far camera plane. 
Both density $\sigma(\mathbf{x})$ and colour $\mathbf{c}(\mathbf{x})$ are modelled using MLPs, and the final rendering is trained in a self-supervised manner using the observed ground-truth per-pixel colours.

To better capture small details, NeRF uses \emph{hierarchical volume sampling} for  $\{t_i\}_{i=1}^{N_\mathrm{depth}}$, 
\textit{i.e.,} 
instead of performing a single rendering pass, points are first sampled in stratified manner. 
The densities at these points are then used for importance sampling in the final pass.
The final model is thus learnt by supervising the rendered pixel colours of both passes with the ground-truth colours. 

\subsection{Spherical Harmonics NeRF}\label{ssec:NeRF_sph_harmonics} 
While \eqref{eq:rendereq} allows for high-quality free viewpoint  synthesis, $\mathbf{c}(\mathbf{x})$ are defined only through an MLP that does not encode the lighting. 
In other words, such formulation learns a Lambertian model of the scene under a fixed lighting. 
The more generalised model with view direction dependencies~\cite{Mildenhall20nerf} learns a slice of the apparent BRDF at a fixed illumination. 
Nonetheless, this learnt representation still does not have a semantic meaning of the underlying scene intrinsics and has no direct control over the lighting. 

To allow relighting, we introduce an explicit 2nd-order Spherical Harmonics (SH) illumination model~\cite{basri2003lambertian} and redefine the rendering equation~\eqref{eq:rendereq} as follows: 
\begin{equation}
    \label{eq:shrendereq}
    \mathbf{C}\left(\{\mathbf{x}_i\}_{i=1}^{N_\mathrm{depth}},\textbf{L}\right)=
    \mathbf{A}\left(\{\mathbf{x}_i\}_{i=1}^{N_\mathrm{depth}}\right)\odot \mathbf{L}\mathbf{b}\left(\mathbf{N}\left(\{\mathbf{x}_i\}_{i=1}^{N_\mathrm{depth}}\right)\right),
\end{equation}
where $\odot$ denotes elementwise multiplication. 
$\mathbf{A}(\mathbf{x})\in\mathbb{R}^3$ is the accumulated albedo colour, generated in the similar way as in \eqref{eq:rendereq}, \textit{i.e.,} by integrating the output of an albedo MLP. 
$\mathbf{L} \in \mathbb{R}^{9\times 3}$ is the per-image learnable SH coefficients, and  $\mathbf{b}(\mathbf{n}) \in \mathbb{R}^9$ is the SH basis. 
$\mathbf{N}(\mathbf{x})$ is the surface normal computed from the accumulated ray density. It is defined as
\begin{equation}
    \mathbf{N}\left(\{\mathbf{x}_i\}_{i=1}^{N_\mathrm{depth}}\right) = \frac{\mathbf{\hat{N}}\left(\{\mathbf{x}_i\}_{i=1}^{N_\mathrm{depth}}\right)}{\left\|\mathbf{\hat{N}}\left(\{\mathbf{x}_i\}_{i=1}^{N_\mathrm{depth}}\right)\right\|^2},
\end{equation}
{\small
\begin{equation}
    \textrm{where }
    \mathbf{\hat{N}}\left(\{\mathbf{x}_i\}_{i=1}^{N_\mathrm{depth}}\right) =
    \sum_{i=1}^{N_\mathrm{depth}}\left(\frac{\partial}{\partial \mathbf{x}_i}\sigma(\mathbf{x}_i)\right)\odot T(t_i)\alpha(\sigma(\mathbf{x}_i)\delta_i).\\
\end{equation}
}
To extract $\mathbf{N}$, we first differentiate the  density of points on the ray with respect to the original $x$-, $y$-, $z$-components of the ray samples, accumulate them over all $N_\mathrm{depth}$ samples on the ray with weights $T(t_i)\alpha(\sigma(\mathbf{x}_i)\delta_i)$, and normalise the resulting vector to a unit sphere. 
Note that in \eqref{eq:shrendereq}, we render in screen space using screen space albedo and normals accumulated from the neural volume. 
The accumulation makes the albedo and surface normal estimates less noisy and aids convergence. It also means we only make a single shading calculation rather than the alternative of one per sample point and accumulating shaded colours.

All terms of \eqref{eq:shrendereq} are learnable except for the SH basis $\mathbf{b}(\cdot)$ and the normal extraction operator $\mathbf{N}(\cdot)$, which are based on fixed, explicit models. 
The proposed lighting model integration allows for explicit relighting by varying 
$\mathbf{L}$.
While it accounts for Lambertian effects, it lacks direct shadow generation, which is crucial for modelling and subsequent relighting of outdoor scenes. 

\subsection{Shadow Generation Network}\label{ssec:shadow_network} 

To allow for explicit shadow control during relighting, we introduce a dedicated shadow model $S\left(\{\mathbf{x}_i\}_{i=1}^{N_\mathrm{depth}}, \mathbf{L}\right)$ and extend the rendering equation  \eqref{eq:shrendereq}: 
{\small
\begin{equation}
    \mathbf{C}\left(\{\mathbf{x}_i\}_{i=1}^{N_\mathrm{depth}},\mathbf{L}\right)=
    S\left(\{\mathbf{x}_i\}_{i=1}^{N_\mathrm{depth}}, \mathbf{L}\right)
    \mathbf{A}\left(\{\mathbf{x}_i\}_{i=1}^{N_\mathrm{depth}}\right)\odot  \mathbf{L}\mathbf{b}\left(\mathbf{N}\left(\{\mathbf{x}_i\}_{i=1}^{N_\mathrm{depth}}\right)\right).
    \label{eq:newrendereq}
\end{equation}
}
The shadow model is defined with a scalar computed by an MLP $s(\mathbf{x}, \mathbf{L}) \in [0, 1]$.
The final shadow value is computed by accumulating along the ray into $S\left(\{\mathbf{x}_i\}_{i=1}^{N_\mathrm{depth}}, \mathbf{L}\right)\in [0,1]$, in the same way as in \eqref{eq:rendereq}. 
Fig.~\ref{fig:pipeline} shows the high-level diagram of the proposed NeRF-OSR. 
Note that the shadow prediction network takes as input the SH coefficients in their  grey-scale version, \textit{i.e.,} $\textbf{L} \in\mathbb{R}^{1\times9}$ and not $\mathbb{R}^{3\times9}$. 
This is motivated by the fact that shadows depend only on the spatial light distribution. 
Unlike traditional ray-tracing approaches as the one used in \cite{Srinivasan21NeRV,Philip19}, our shadow estimator operates much more efficiently, through just same single forward pass as albedo and geometry. 
We argue that it is a strength, as it makes the method much more computationally scalable, while still allowing us to relight using completely new illumination conditions. %

\subsection{Objective Function}\label{ssec:objective_function} 

We optimise the following loss function: 
\begin{equation}\label{eq:costfunction}
    \mathcal{L}(\mathbf{C}, \mathbf{C}^\mathrm{(GT)}, S)=
    \mathrm{MSE}(\mathbf{C}, \mathbf{C}^\mathrm{(GT)})+
    \lambda\,\mathrm{MSE}(S, 1),
\end{equation} 
where $\mathrm{MSE}(\cdot, \cdot)$ is the mean squared error. 
The first term is a reconstruction loss defined on the estimated colour $\mathbf{C}$ and the corresponding ground truth $\mathbf{C}^\mathrm{(GT)}$. 
The second term regularises shadows. 
The shadow network $S$ absorbs all greyscale lighting effects that cannot be explained by SH.
To limit it to learning only shadows, we 
select the largest value of the regularisation strength $\lambda$ that does not degrade the PSNR of the reconstructed images.
Experimentation shows that removing the regulariser usually leads to $S$ 
learning all the illumination components, except for the chromaticity---thus making the SH lighting useless.

\subsection{Training and Implementation Details}\label{ssec:training_details} 
Our self-supervised model is trained on RGB images of an outdoor scene photographed from various viewpoints and under different illumination. 
We next describe several strategies for training our method and their importance. 

\textbf{Frequency Annealing.}
We noticed empirically that training the  model as-is leads to noisy normal maps. 
Above some threshold on the number of the positional encoding (PE) frequencies, the initially generated noise (at the start of the training) becomes very hard to manipulate; it hardly converges to the correct geometry. 
Hence, we alleviate this by using the annealing scheme slightly modified from Deformable NeRF~\cite{park2020nerfies}, \textit{i.e.,} we add an annealing coefficient $\beta_k(n)$ to each of the PE components $\mathbf{\gamma}_k(\mathbf{x})$: 
$\mathbf{\gamma}_k'(\mathbf{x})=\mathbf{\gamma}_k(\mathbf{x})\beta_k(n)$, where
$\beta_k(n)=\frac{1}{2}(1-\cos\left(\pi\mathrm{clamp}(\alpha-i+N_\mathrm{fmin},0, 1)\right))$, $\alpha(n)=(N_\mathrm{fmax}-N_\mathrm{fmin })\frac{n}{N_\mathrm{anneal}}$, $n$ is the current training iteration, $N_\mathrm{fmax}$ is the total number of used PE frequencies (the proposed model uses $12$), $N_\mathrm{fmin}$ is the number of used PE frequences at the start (we use $8$), $N_\mathrm{anneal}$ is tuned empirically to $3\cdot 10^{4}$ for all sequences. 
This training strategy enables significantly  improved geometry predictions. 

\textbf{Ray Direction Jitter.}
To improve the generalisability of NeRF-OSR, we apply a sub-pixel jitter to the ray direction. 
Here, instead of shooting in the pixel centres, a jitter 
$\mathbf{\psi}$ is used as follows: 
$x_i=\mathbf{o}+t_i(\mathbf{d}+\mathbf{\psi})$. We sample $\mathbf{\psi}$ uniformly,  
such that the resulting ray still confines to the boundaries of its designated pixel.

\textbf{Shadow Network Input Jitter.} 
Since the shadows are generated in a learning-based fashion instead of using direct geometric approaches, there remains the possibility of overfitting to the training lightings. 
To mitigate this effect, we add a slight normal noise $\varepsilon$ to the environment coefficients as input of the shadow generation network: 
\begin{equation}\label{eq:normal_noise} 
    S'\left(\{\mathbf{x}_i\}_{i=1}^{N_\mathrm{depth}}, \mathbf{L}\right)=
    S\left(\{\mathbf{x}_i\}_{i=1}^{N_\mathrm{depth}}, \mathbf{L}+\mathbf{\varepsilon}\right),
\end{equation}
where $\mathbf{\varepsilon}{\sim}\mathcal{N}(0, 0.025I)$.
\eqref{eq:normal_noise} can be interpreted as a locality condition, \textit{i.e.,} in similar lighting conditions, shadows should not be too different. 
This allows the model to learn smoother transitions  between different lightings.

\textbf{Implementation.} 
We use NeRF++~\cite{KaiZhang2020} with the background network disabled as the code base
and work within the unit sphere bounds of the foreground network. 
For training and evaluation, we use two Nvidia Quadro RTX 8000 GPUs. 
We train the model for $5 \cdot 10^5$ iterations using a batch size of $2^{10}$ rays, which takes ${\approx}2$ days. 

%% file: figures/pipeline.tex
\begin{figure}[!t]
    \centering
    \includegraphics[width=1.0\linewidth]{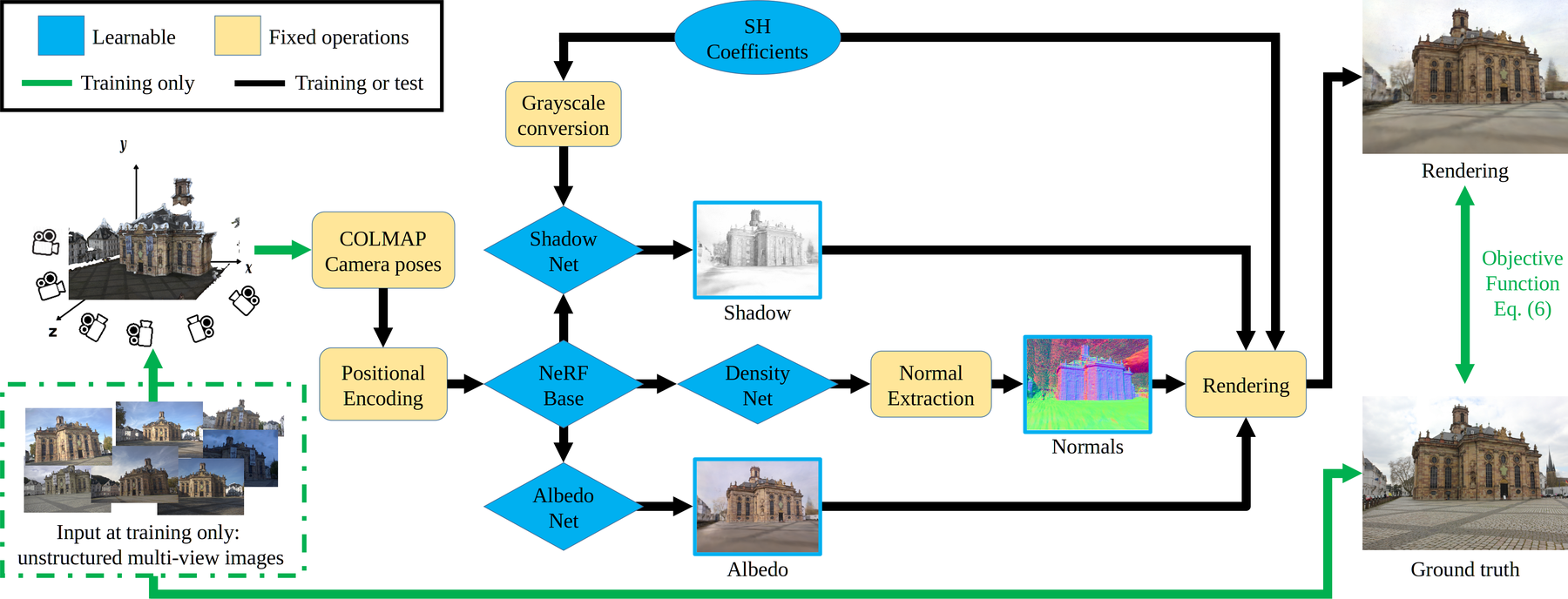}
\caption{\textbf{Our NeRF-OSR uses outdoor images of a site photographed in an uncontrolled setting (dashed green) to recover a relightable implicit scene model.} 
It learns the scene intrinsics and illumination as expressed by the SH coefficients. 
Here, a dedicated neural component learns shadows. 
During the test, NeRF-OSR can synthesise novel images at arbitrary camera viewpoints and scene illumination; the user directly supplies the desired camera pose and the scene illumination, either from an environment map or via SH coefficients.
} 
    \label{fig:pipeline}
\end{figure}

%% file: 4_dataset.tex
\section{A New Benchmark for Outdoor Scene Relighting} 
\label{sec:dataset} 

\input{figures/dataset} 
\input{figures/envmap}

Several datasets for outdoor sites  exist~\cite{MegaDepthLi18,Li20crowdsampling,Snavely06,Jin2020,Yu20}. 
Most of them~\cite{MegaDepthLi18,Snavely06,Jin2020,Li20crowdsampling} were collected with the task of 3D scene reconstruction in mind and not relighting. 
Hence, they mostly contain publicly available photos collected in uncontrolled settings. 
Furthermore, they do not provide environment maps, which are important for evaluating relighting techniques numerically on real data against ground truth. 
Examples of such datasets are the PhotoTourism~\cite{Snavely06,Jin2020} and the MegaDepth~\cite{MegaDepthLi18}. 
The MegaDepth dataset consists of multi-view images of several sites that were initially a part of the Landmarks10k dataset~\cite{Li12}. 
Here, the depth signal is extracted using  COLMAP~\cite{Schonberger16colmap} and the multi-view stereo (MVS) approach  \cite{Schonberger16}. 
While MegaDepth was originally released as a benchmark for single-view depth extraction, it was used by Yu~\etal~\cite{Yu20} (one of the most recent relighting works). However, it can only evaluate methods qualitatively.

To allow for numerical evaluation on real data against ground truth, Yu~\etal~\cite{Yu20}
recorded \textit{one site} from different viewpoints and at different times of the day using a DSLR camera, along with the environment maps. 
Unfortunately, this benchmark is limited in two ways: 
First, it contains a single site. 
Second, the captured environment maps were not colour-corrected with respect to the DSLR camera of the main recordings. 
\textit{Hence, numerical results obtained with this dataset would always differ from the ground truth by an unknown, possibly nonlinear, colour transformation.} 
Therefore, any error metric must first compute an optimal transformation (Yu~\etal~\cite{Yu20} used a per-colour channel linear scaling). 
This makes it hard to separate the behaviour of the examined relighting methods from the corrective behaviour of this normalisation. 

Hence, we present a new benchmark for outdoor scene relighting. 
Our dataset is the first of its kind in terms of size and the ability to perform accurate numerical evaluations on real data against ground truth. 
It is much larger than Yu~\etal~\cite{Yu20}, containing eight sites captured from various viewpoints using a DSLR camera (3240 viewpoints captured in 110 different recording sessions). 
Multiple recording sessions were performed for each site, at different times of the day; all sessions cover different weathers, including sunny and cloudy days. 
We also capture a $360^\circ$ shot of the environment map for each session. 
Unlike Yu~\etal~\cite{Yu20}, we explicitly account for the colour calibration between the environment maps and the DSLR camera of the main recordings. 
To this end---for every session in the test set---we also simultaneously capture the ``GretagMacbeth ColorChecker'' colour calibration chart with the DSLR and the $360^\circ$ cameras. 
We then apply the second-order method of Finlayson~\etal~\cite{Finlayson15} to colour-correct the environment maps by calibrating their ColorChecker values to the ColorChecker values of the corresponding DSLR image. 
Finally, we manually align the environment maps to the world coordinates using COLMAP~\cite{Schonberger16colmap} reconstructions of each site. 

Fig.~\ref{fig:dataset} shows samples from the various sites from our dataset and the  corresponding environment maps.  
See Fig.~\ref{fig:envmap} for the  colour-corrected environment maps. 
Note that we target scenes with minimal specular effects  (such as brick or wood buildings). 
All data was captured in exposure brackets of five photos ranging from -3 to +3 EV for the DSLR photos and from -2 to +2 EV for the environment maps. 
We used the darkest capture for the $360^\circ$ environment maps so that the sun is least overexposed. 
For the ColorChecker calibration with DSLR, we use images that are dark enough so that the white cells of the chequerboard are not overexposed. 
The DSLR image resolution is $5184{\times}3456$ pixel, while the resolution of the environment maps is $5660{\times}2830$ pixel.

%% file: figures/dataset.tex
\begin{figure}[!t]
    \centering
    \includegraphics[width=1.0\textwidth, trim=0 0.9cm 0 0, clip]{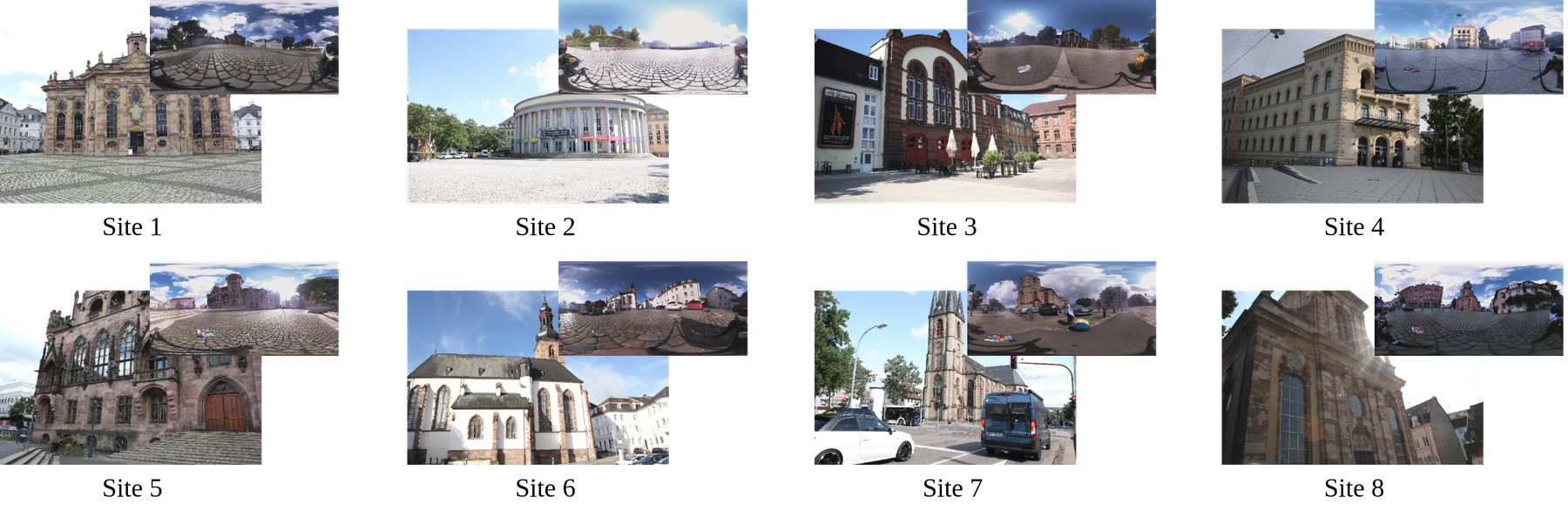}
\caption{Sample views from the new benchmark dataset for outdoor scene relighting.
The dataset has 3240 views captured in 110 different recording sessions.
}
    \label{fig:dataset}
\end{figure}

%% file: figures/envmap.tex
\begin{figure}[!t]
    \centering
    \includegraphics[width=1.0\linewidth]{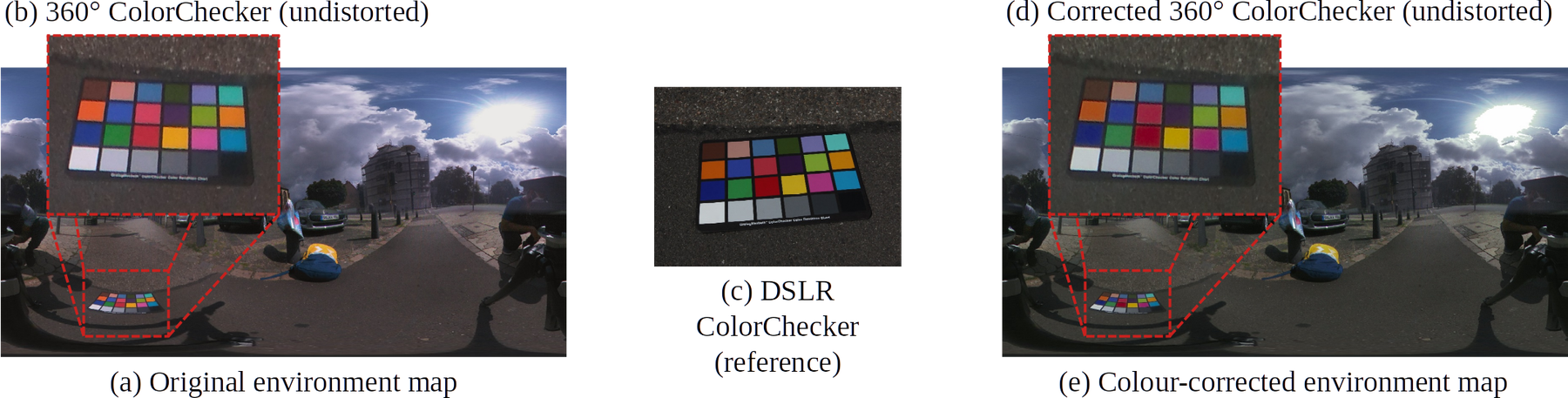}
\caption{For each recording session, we capture a colour chequerboard with the DSLR and $360^\circ$ cameras. 
We colour-correct the $360^\circ$ maps to match the DSLR.
}%
    \label{fig:envmap}
\end{figure}

%% file: 5_results_ye.tex
\section{Results}

We evaluate the performance of NeRF-OSR on various real-world sites. 
We examine three sites from our newly proposed dataset and the {Trevi Fountain} from the PhotoTourism  dataset~\cite{Brualla20nerfw}. 
These scenes include a variety of features. This includes large and small scale details as the sculptures (Site 1, Trevi), structural details such as trees and
umbrellas (Site 3), a piecewise-smooth surfaces casting a lot of shadows on itself (Site 2), water (Trevi) and surrounding buildings (in all). Furthermore, {Trevi Fountain} shows performance
on data collected completely from the internet through crowdsourcing. 
Note that only qualitative evaluation on {Trevi Fountain} is possible due to the absence of environment maps.
We also evaluate the various design choices of our method in an ablative study. 

Among existing scene relighting methods (see Sec.~\ref{sec:related}), 
we primarily compare against 
Yu~\etal~\cite{Yu20} and  Philip~\etal\cite{Philip19} as they handle a similar type of input data like ours; outdoor scenes photographed in uncontrolled settings and have a direct semantic understanding of the scene illumination. 
We note however, despite this, both Yu~\etal~\cite{Yu20} and  Philip~\etal\cite{Philip19} are designed to edit only the illumination of the input image, while our method can edit both the illumination and the viewpoint. 
This makes both these methods \cite{Yu20,Philip19} not direct competitors
to our method, but still the most related in literature.
NeRV~\cite{Srinivasan21NeRV} can not be
applied to our data as their setup is fundamentally different from ours. It requires a training scene to be illuminated by known lighting while our technique uses data shot
in unknown lightings.
We also do not compare quantitatively against NeRF-W \cite{Brualla20nerfw} or other style-based based methods as they do not perform intrinsic decomposition, don't have a physical understanding of the scene illumination and can not edit lighting according to an environment map. 
In contrast, the intrinsic decomposition of NeRF-OSR enables applications that are inaccessible for style-based methods. 
For instance, we show how we can edit the albedo of an examined scene and relighting the entire resulting composited scene (Fig.~\ref{fig:vrAlbedoEditing}-middle).
We also show how our method can achieve real-time rendering with conventional computer graphics methods using the extracted mesh and albedo (Fig.~\ref{fig:vrAlbedoEditing}-left).

We note that Boss~\etal~\cite{Boss20NeRD} (NeRD) requires the examined object to be at a similar distance from all views---an assumption that is fundamentally violated for outdoor data captured in uncontrolled setup and in our data.
As confirmed by the authors of NeRD, this makes the reconstruction nearly impossible for our data.
Furthermore, attempting to run NeRD on our data by the paper authors resulted in ray distance variation that is very large, requiring a large number of samples per ray. Thus it was computationally infeasible to process our data with NeRD.
NeRF-OSR is the first method that can simultaneously edit the viewpoint and lighting of outdoor sites using neural radiance fields.
It also extracts the underlying scene intrinsics and has a dedicated illumination-based shadow component 
It produces photorealistic results and significantly outperforms state of the art. 
It is also not limited by synthesising soft shadows only and can synthesise novel hard shadows as well (see Figs.~\ref{fig:teaser} and \ref{fig:viewpointedit}).

\input{figures/viewpointedit}

\textbf{Data Pre-Processing.}
Since NeRF-OSR does not aim to synthesise dynamic objects and discards them 
(\textit{e.g.,} cars, people and bikes) from the training stage. 
Although we attempted to reduce their presence during our recordings, the uncontrolled nature of the data makes eliminating them during capture impossible. 
We, therefore, use the segmentation method of Tao~\etal~\cite{Tao20} to obtain high-quality masks of such objects. 
Furthermore, even though NeRF-OSR can synthesise the sky and vegetation (\textit{e.g.,} trees), it is not possible to evaluate their predictions due to their highly varying appearance, especially when recordings sessions span different weather seasons. 
Hence, we also estimate the masks of these regions and exclude them from our evaluation. 
For Sites 1--3, we keep five recording sessions for testing and use the rest for training. 
The resulting training/test splits are: 160/95 views for {Site 1}, 301/96 views for {Site 2} and 258/96 views for {Site 3}.

\textbf{Relighting with Ground-Truth Environments.}
We quantitatively evaluate the parametric lighting control of our method and show that it can reproduce novel lighting using lighting coefficients extracted from environment maps. 
From each recording session of our dataset, we select one photo from the test set as the source. 
With Site 1, this gives five source images in total. 
We render all five images at the observed viewpoints and illumination directly. 
However, for Philip~\etal~\cite{Philip19} and Yu~\etal~\cite{Yu20},  only the illumination of a given image can be edited. 
Hence, for each source image, we relight it using the illumination of the four other source images of the same site.  
We then cross-project the output to the camera viewpoint from which the target illumination was extracted. 
This is done by utilising the COLMAP reconstructions. 
We use segmentation
masks to evaluate performance on regions where consistent predictions can be made and cross-projected for other methods. This is usually the main building. Here, we compute several metrics, including MSE, MAE and SSIM. For SSIM we report the average over the segmentation mask, using scikit-image~\cite{scikitimage} implementation. Here, we use an SSIM metric with a window size of 5 and the segmentation mask eroded by the same window size to remove the impact of the pixels outside of the mask on the metric value.

Tab.~\ref{tbl:numrelighting} reports the results of this experiment (the averages over all evaluated images). 
Fig.~\ref{fig:relightingGT} shows several  ground-truth images and views rendered by the compared methods. 
NeRF-OSR outperforms related techniques quantitatively and qualitatively.
While our method and Philip~\etal generate results at  $1280{\times}844$ pixel, Yu~\etal~can only generate results at  $303{\times}200$ pixel. 
Hence in Tab.~\ref{tbl:numrelighting} we also compare against Yu~\etal~in a setting where we downscale the output of our method to Yu~\etal's default resolution (see d/s in Tab.~\ref{tbl:numrelighting}).
Despite this, our method still outperforms Yu~\etal which shows our more superior performance is not due to differences in the output resolution. 
We note that comparing against style-based methods like NeRF-W here is not feasible as they do not have a semantic representation of light and thus can not edit the light according to an environment map. 
\input{results_table}

\input{figures/relightingGTYU}

\textbf{Ablation Study.}
\label{sec:ablation}
We evaluate the design choices of our method through an ablation study. 
We follow the same evaluation procedure as for the relighting comparison and report results as an average taken over all output images. 
For our approach, that are five images of Site 1. 
For Philip~\etal~\cite{Philip19}, that are 20 images in total.
Tab.~\ref{tbl:numrelighting}-(bottom) reports the PSNR, MSE, MAE and SSIM of various tested settings. 
Results show that the best performance is obtained by using the full version of NeRF-OSR. We note that since all metrics in Tab.~\ref{tbl:numrelighting} are computed only over masked regions, they are expected to be of a higher performance if the entire image was evaluated, while filling the unmasked
regions with black.

\textbf{Real-time Interactive Rendering in VR.} %
In contrast to style-based methods such as \cite{Brualla20nerfw}, our rendering is an explicit function of geometry, albedo, shadow, and the lighting conditions (see Eq.~\ref{eq:newrendereq}).
Our model provides direct access to albedo and geometry.
The lighting and shadows can be generated from the geometry using multiple, potentially non-differentiable, lighting models at render-time.
Hence, if we can extract geometry and albedo at sufficient resolution, we can use them without the slow NeRF ray-marching at little to no loss of quality, compared to the original neural model.

We extract the geometry and albedo from the learned model of Site 1 as a mesh using Marching Cubes~\cite{marchingcubes} at resolution  
$1000^3$ 
voxels. 
Then we use them in our interactive VR renderer implemented with C++, OpenGL and SteamVR. 
The lighting model consists of the sun and a simple geometry-based shadow map~\cite{shadowmap}:
$
    \mathbf{C}_\mathrm{interactive}=\mathbf{C}_\mathrm{ambient}+s \odot \mathbf{C}_\mathrm{sun} \max\{0, \mathbf{D}^T_\mathrm{sun} \mathbf{N}\},
$ where $\mathbf{C}_\mathrm{ambient}$ is the ambient colour, $s$ is 0 when the rendered point is occluded and 1 if not, according to the shadow map, $\mathbf{C}_\mathrm{sun}$ is the colour of the sun, $\mathbf{D}_\mathrm{sun}$ is the direction of the sun and $\mathbf{N}$ is the normal of the mesh.
The user can interactively move in the scene and control the sun direction with their controllers.
The demo runs in real-time on a desktop computer with an Intel i7-4770 CPU, an Nvidia GeForce GTX 970 (4GB VRAM) GPU and an Oculus Rift S HMD.
The system RAM usage of the application is below 3 GB.
We provide an extensive demo in the supplementary video and show an extract in Fig.~\ref{fig:vrAlbedoEditing}-(left).
\input{figures/vralbedoediting}

\textbf{Albedo and Shadow Editing.} %
Another application of our intrinsic decomposition is to edit the scene albedo, without affecting the illumination or shadows.
Such application is not possible by style-based methods by any means, \eg, NeRF-W~\cite{Brualla20nerfw}, as they do not perform image decomposition.
In Fig.~\ref{fig:vrAlbedoEditing}-(middle), we replace the announcement poster in Site 3 with an ECCV 2022 poster.
Note how the replaced poster looks natural with the rest of the scene.
In Fig.~\ref{fig:vrAlbedoEditing}-(right), we edit the shadow strength post-render.
Please find extended video results of this experiment in our supplementary video, where we also show relighting results with the composited announcement poster.

%% file: figures/viewpointedit.tex
\begin{figure}[!t]
    \centering
    \includegraphics[width=1.0\linewidth]{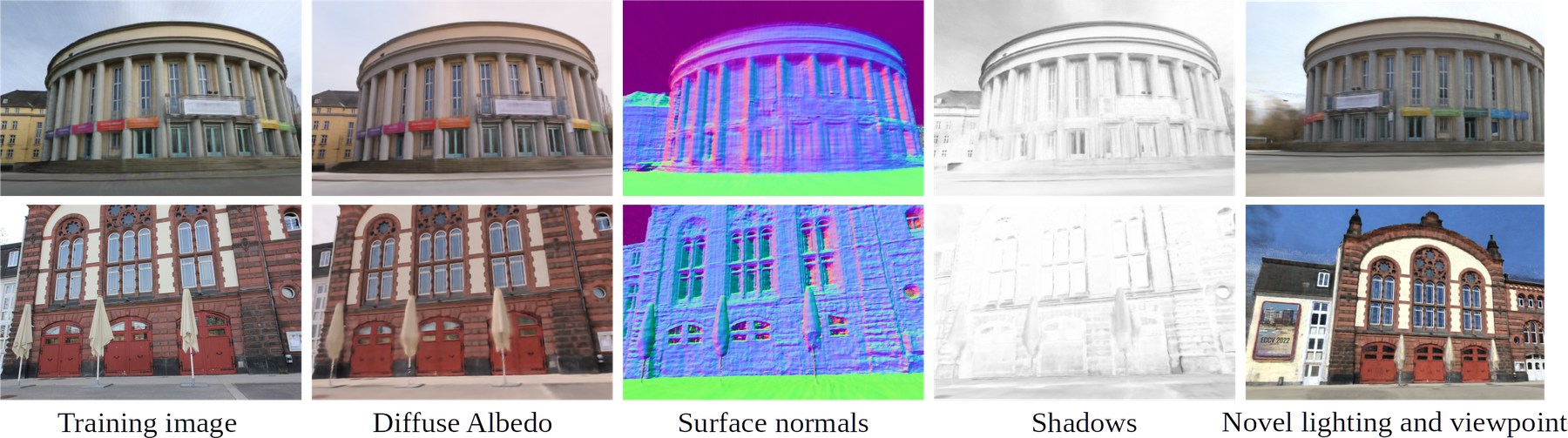}
\caption{Our NeRF-OSR renders photorealistic novel views and simultaneously edits lighting. It also estimates the underlying scene semantics including a dedicated shadows component. Moreover, it can also synthesise hard shadows.%
}
    \label{fig:viewpointedit}
\end{figure}

%% file: results_table.tex
\begin{table}[!t]
    \centering

    \resizebox{6cm}{!}{
    \begin{tabular}{c|c|c|c|c}
         Method & PSNR $\uparrow$ & MSE $\downarrow$ & MAE $\downarrow$ & SSIM $\uparrow$ \\
         \hline
         &\multicolumn{4}{c}{Site 1}\\
         \hline
         Yu~\etal~\cite{Yu20} & $18.71$ & $0.014$  & $0.088$ & $0.4~~~$ \\
         Philip~\etal~\cite{Philip19} (d/s) & $17.37$ & $0.019$ & $0.105$ & $0.429$\\
         Ours (d/s) & $\mathbf{19.86}$ & $\mathbf{0.011}$ & $\mathbf{0.08~}$ & $\mathbf{0.626}$ \\
         \hline
         Yu~\etal~\cite{Yu20} (u/s) & $17.87$ &  $0.017$ & $0.097$ & $0.378$ \\
         Philip~\etal~\cite{Philip19} & $16.63$ & $0.023$  & $0.113$  & $0.367$ \\
         Ours & $\mathbf{18.72}$ & $\mathbf{0.014}$  & $\mathbf{0.09~}$ & $\mathbf{0.468}$ \\
         \hline
         No shadows & $17.82$ & $0.017$  & $0.101$  & $0.418$ \\
         No annealing & $17.16$ & $0.02~\,$  & $0.108$   & $0.324$ \\
         No ray jitter & $18.43$ & $0.015$  & $0.093$   & $0.433$ \\
         No shadow jitter & $18.28$ & $0.016$  & $0.095$  & $0.413$ \\
         No shadow regulariser & $17.62$ & $0.018$  & $0.105$ & $0.373$ \\
         \multicolumn{5}{c}{}\\
         \multicolumn{5}{c}{}\\
         \end{tabular}}
         \resizebox{6cm}{!}{
         \begin{tabular}{c|c|c|c|c}
         Method & PSNR $\uparrow$ & MSE $\downarrow$ & MAE $\downarrow$ & SSIM $\uparrow$ \\
         \hline
         &\multicolumn{4}{c}{Site 2}\\
         \hline
         Yu~\etal~\cite{Yu20} & $15.43$ & $0.031$  & $0.136$ & $0.363$  \\
         Philip~\etal~\cite{Philip19} (d/s) & $11.85$ & $0.07~$ & $0.21~$ & $0.184$  \\
         Ours (d/s) & $\mathbf{15.83}$ & $\mathbf{0.026}$ & $\mathbf{0.128}$ & $\mathbf{0.556}$  \\
         \hline
         Yu~\etal~\cite{Yu20} (u/s) & $15.28$ &  $0.032$ & $0.138$ & $0.385$  \\
         Philip~\etal~\cite{Philip19} & $12.34$ & $0.065$  & $0.2~~~$  & $0.272$  \\
         Ours & $\mathbf{15.43}$ & $\mathbf{0.029}$  & $\mathbf{0.133}$ & $\mathbf{0.517}$  \\
         \hline
         &\multicolumn{4}{c}{Site 3} \\
         \hline
         Yu~\etal~\cite{Yu20} & $15.84$ & $0.028$ & $0.123$ & $0.392$ \\
         Philip~\etal~\cite{Philip19} (d/s)  & $12.85$ & $0.054$ & $0.169$ & $0.164$ \\
         Ours (d/s) & $\mathbf{17.38}$ & $\mathbf{0.021}$ & $\mathbf{0.106}$ & $\mathbf{0.576}$ \\
         \hline
         Yu~\etal~\cite{Yu20} (u/s) & $15.17$ & $0.033$ & $0.133$ & $0.376$ \\
         Philip~\etal~\cite{Philip19}  & $12.28$ & $0.062$ & $0.179$ & $0.319$ \\
         Ours &  $\mathbf{16.65}$ & $\mathbf{0.024}$ & $\mathbf{0.114}$ & $\mathbf{0.501}$ \\
    \end{tabular}}
    \caption{Quantitative evaluation of the relighting capabilities of different techniques. 
    We report the metrics for 
    Sites 1, 2 and 3 from our dataset. 
    Our technique significantly outperforms related methods~\cite{Yu20,Philip19}. ``d/s'' and ``u/s'' are shorthands for ``downscaled'' and ``upscaled'', respectively. 
    Bottom left: ablation study of our various design choices. 
    Our full model achieves the best result.
    }
    \label{tbl:numrelighting}
\end{table}

%% file: figures/relightingGTYU.tex
\begin{figure}[!t]
    \centering
    \includegraphics[width=0.495\linewidth,clip,trim={0cm 1098px 0cm 0px}]{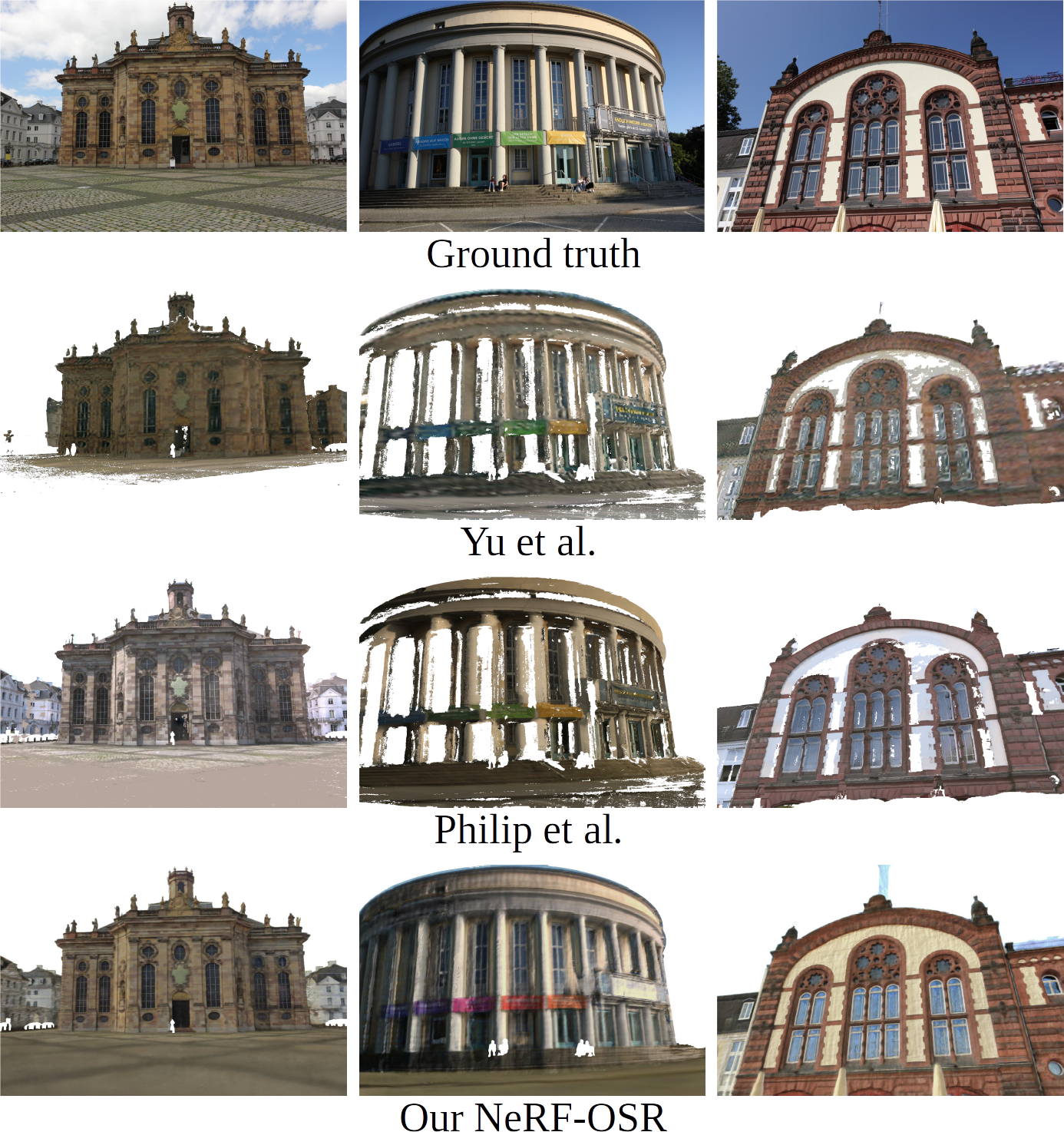} %
    \includegraphics[width=0.495\linewidth,clip,trim={0cm 368px 0cm 745px}]{figures/gtrelighting_aligned2_wide.png} %
    \includegraphics[width=0.495\linewidth,clip,trim={0cm 0px 0cm 1112px}]{figures/gtrelighting_aligned2_wide.png} %
    \includegraphics[width=0.495\linewidth,clip,trim={0cm 737px 0cm 360px}]{figures/gtrelighting_aligned2_wide.png} %
    
\caption{Relighting using ground-truth environment map. Since Philip~\etal~\cite{Philip19} and Yu~\etal~\cite{Yu20} can not edit the camera viewpoint---unlike NeRF-OSR---we cross-project their result on the ground-truth view. 
Our approach captures the illumination significantly better than related methods. 
See Tab.~\ref{tbl:numrelighting} for the corresponding numerical evaluations.%
}
    \label{fig:relightingGT}
\end{figure}

%% file: figures/vralbedoediting.tex
\begin{figure}[!t]
    \centering
    
        \includegraphics[width=0.32\textwidth, trim=140px 80px 200px 110px, clip]{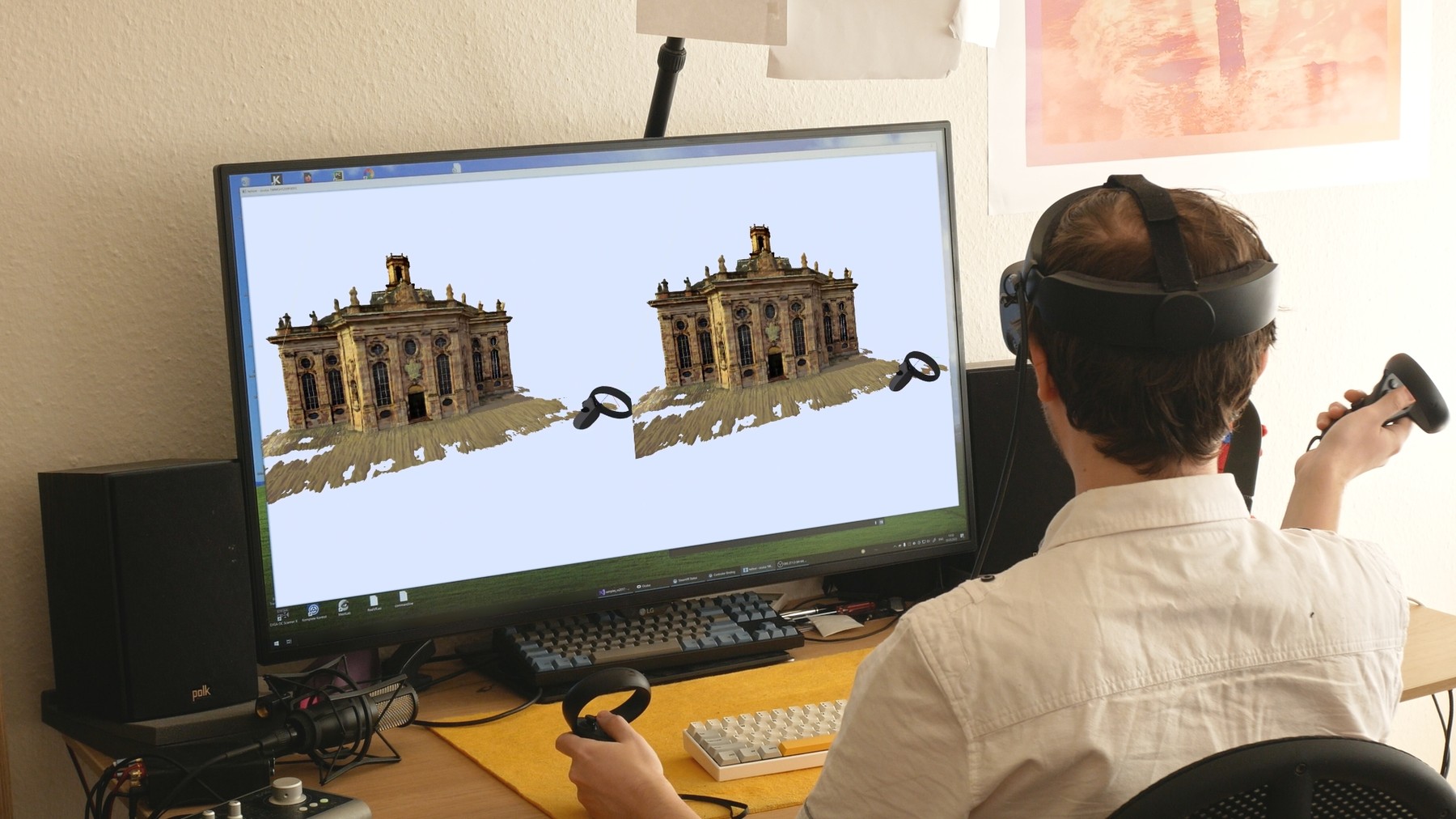}
         \includegraphics[width=0.28\textwidth, trim=0px 0cm 0 0, clip]{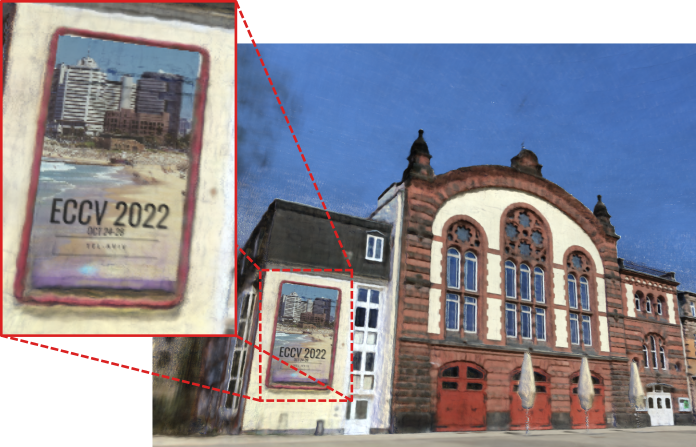}
         \includegraphics[width=0.36\textwidth, trim=0px 0cm 0 0, clip]{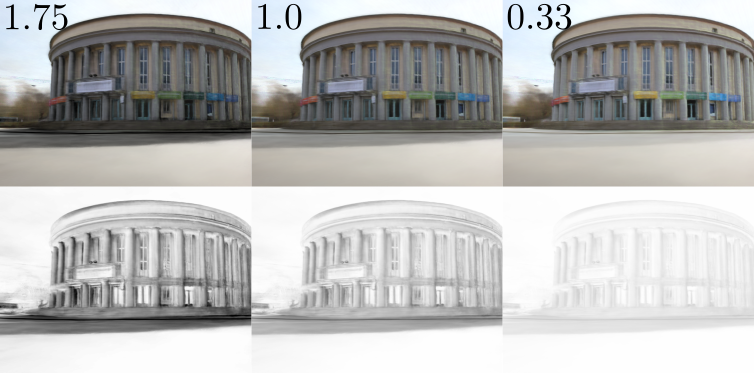}
\caption{(Left) Real-time interactive rendering of the extracted model in VR (screen capture is overlayed over the display for clearer image). Here, the sunlight illuminates the left side of the building and casts clear hard shadows on the right side. And an example of editing the scene albedo (middle) and shadows (right), independently of illumination and other intrinsics.
}
    \label{fig:vrAlbedoEditing}
\end{figure}

%% file: 6_discussion.tex
\section{Discussion and Conclusion} 
We have shown that the second-order SH lighting model is capable of producing plausible relightings. 
While sunlit environments can contain shadows not well represented by a second-order SH, we believe our learned shadow component compensates for this (see Figs.~\ref{fig:vrAlbedoEditing}, right, for examples of novel hard shadows). 
Nevertheless, the SH illumination model can still be restricted in terms of high-frequency illumination, specularities and spatially varying illumination. 
Capturing such effects would enable reconstruction of view-dependent effects and more challenging scenes, including nighttime conditions. 

Despite our method outperforms related approaches numerically and visually, some blur could exist. 
We believe this is due to some inaccuracies in geometry estimation. 
More specifically, we learn a disentangled representation of the image intrinsics, allowing many novel applications
(Fig.~\ref{fig:vrAlbedoEditing}). 
This, however, requires precise geometry, as even tiny bumps in the learned geometry can lead to significant change in normals and, hence, errors in the computed illumination.
Hence the model can smooth some parts of the geometry in favour of having more accurate lighting. 
This leads to overall better relighting results, compared to other methods as shown in Tab.~\ref{tbl:numrelighting}. 
Nevertheless, future work can further improve results by examining more sophisticated geometry models (\textit{e.g.,} a hybrid volume density or implicit surface representation \cite{Oechsle2021ICCV,wang2021neus}).
Our method needs only a set of in-the-wild photos taken from different times and views.
To this end, we have evaluated our approach on our newly collected dataset and on the "Trevi" scene from \cite{Jin2020,Snavely06}. 
This scene was collected completely from the Internet and is widely used in literature~\cite{Li20crowdsampling,Brualla20nerfw,Meshry19}. 
Recall that ground-truth environment maps are only used for evaluation (as in Sec.~\ref{sec:ablation}) and not required for our method to work. 
While the datasets we examined show practical use-cases of our method,  future work could investigate using as few as a single illumination condition during test. 
Finally, incorporating more priors of the outdoor scenes could be an interesting future research direction.

%% file: 7_conclusion.tex
\noindent \textbf{Concluding Remarks.} 
We presented the first method for simultaneous novel view and novel lighting generation of outdoor scenes captured from uncontrolled settings. 
We have shown that posed images with varying illumination are sufficient to train a neural representation of scene intrinsics and estimate per-image illumination. 
Our method outperforms related techniques subjectively and quantitatively on several sequences, including the newly collected benchmark dataset with ground-truth environment maps. 

{%
\noindent\textbf{Acknowledgements.} 
We thank Christen Millerdurai for the help with the dataset recording. 
This work was supported by the ERC Consolidator Grant 4DRepLy (770784). 
}

%% file: supplemental.tex
This supplemental document provides more details on the new dataset for outdoor scene relighting, spherical harmonics (SH) environment estimation, ablative study, the experiments, and additional visualisations. 

\section{SH Environment Estimation}

Recall that our method represents the target illumination through SH coefficients. For experiments in Sec.~5 we estimate the SH coefficients from a 360 environment photo using least squares, inline with Yu et al.~\cite{Yu20}. Here, the 360 environment maps allows us to perform numerical evaluations against ground truth (see Sec.~5).
We also show in Sec.~5 interactive relighting application where the SH coefficients are directly controlled by the user. Furthermore, the supplemental video (at 3:31-3:43) shows results under novel lighting using externally defined SH coefficients.

\section{Statistics of the Dataset} 

\input{figures/dataset_tbl}

Tab.~\ref{tbl:dataset_summary} lists the statistics of our new dataset. 
It consists of eight sites photographed from 3240 views in 110 different sessions. 
Our dataset is the first to allow numerical evaluation of relighting methods on real data against ground truth, thanks to the environment maps and the captured colour chequerboards. 
We believe it will be valuable for the community, and we plan to release it.

\section{Ablative Study}

\markboth{\the\authorrunning}{\the\titlerunning}

Fig.~\ref{fig:ablationCol} demonstrates the impact of the design choices in  NeRF-OSR on the final novel view renderings with relighting. 
Not using frequency annealing leads to an evident degradation in the output (the fourth row). 
This includes circular-shaped artefacts on the ground (the second and the third columns) and clear artefacts on the building (the first three columns, from the left). 
Removing the shadow regulariser often causes the shadow layer to learn all the illumination components, except the chromaticity, leading to significant artefacts (the first and the last columns). 
Removing the ray jitter leads to clear artefacts, as shown in the first column. 
Finally, removing shadow learning and shadow jitter produces less accurate reconstruction (the third column). 
The strength of shadow learning is more evident during timelapse relighting (see the supplementary video). 
The full model produces the best results, which is also reflected numerically in Tab.~1 of the main manuscript.

\section{Video Results}

We demonstrate the ability of NeRF-OSR to edit the camera  viewpoint and illumination in our supplementary video.
Thus, we show timelapse relighting, where the camera viewpoint is fixed and the illumination changes by rotating the %
lighting $360^\circ$ around the building. 
Our approach handles known lighting conditions and can generalise well to new ones. 
Even though some synthesised lightings can not occur in real life (\textit{e.g.,} due to the sun trajectory covering only $180^\circ$ of the sky at most), \mbox{NeRF-OSR} still produces a highly photorealistic output. 
We also change the viewpoint while keeping the scene illumination fixed. 
Moreover, we show results when both scene illumination and viewpoint were not seen during the training. 
Finally, we visualise the scene intrinsics, \textit{i.e.,} normals,  albedo, shadow and shading, as recovered by our technique. 

Fig.~\ref{fig:additionalvis} provides several screenshots from the  video results, including simultaneous relighting and novel view synthesis of Site 3  (Fig.~\ref{fig:additionalvis}-(a)). In the figure, we show strong hard shadow synthesis.
Next, NeRF-OSR can be used for relighting using  unrealistic and synthetic lighting  (Fig.~\ref{fig:additionalvis}-(b)). 
Finally, while NeRF-W \cite{Brualla20nerfw} can interpolate between the learnt appearances, our method decomposes the scene in its intrinsic components; it  enables manipulation of the lighting, which results in interpretable editing of the novel views  (Fig.~\ref{fig:additionalvis}-(c))). Also this allows direct editing of albedo, shadows, independently of other intrinsics, real-time interactive VR rendering of the extracted model, which we demonstrate in Secs.~5.4-5.5 of the main manuscript and in the supplementary video. Such shown applications are not possible with style-based techniques by any means as they do not perform intrinsics decomposition.

\section{Additional Renderings of Various Sites}
\begin{figure}
    \centering
    \begin{tabular}{@{}c@{\hspace{0.05cm}}c@{\hspace{0.05cm}}c@{\hspace{0.05cm}}c@{\hspace{0.05cm}}c@{}}
       \includegraphics[height=1.6cm,clip=true]{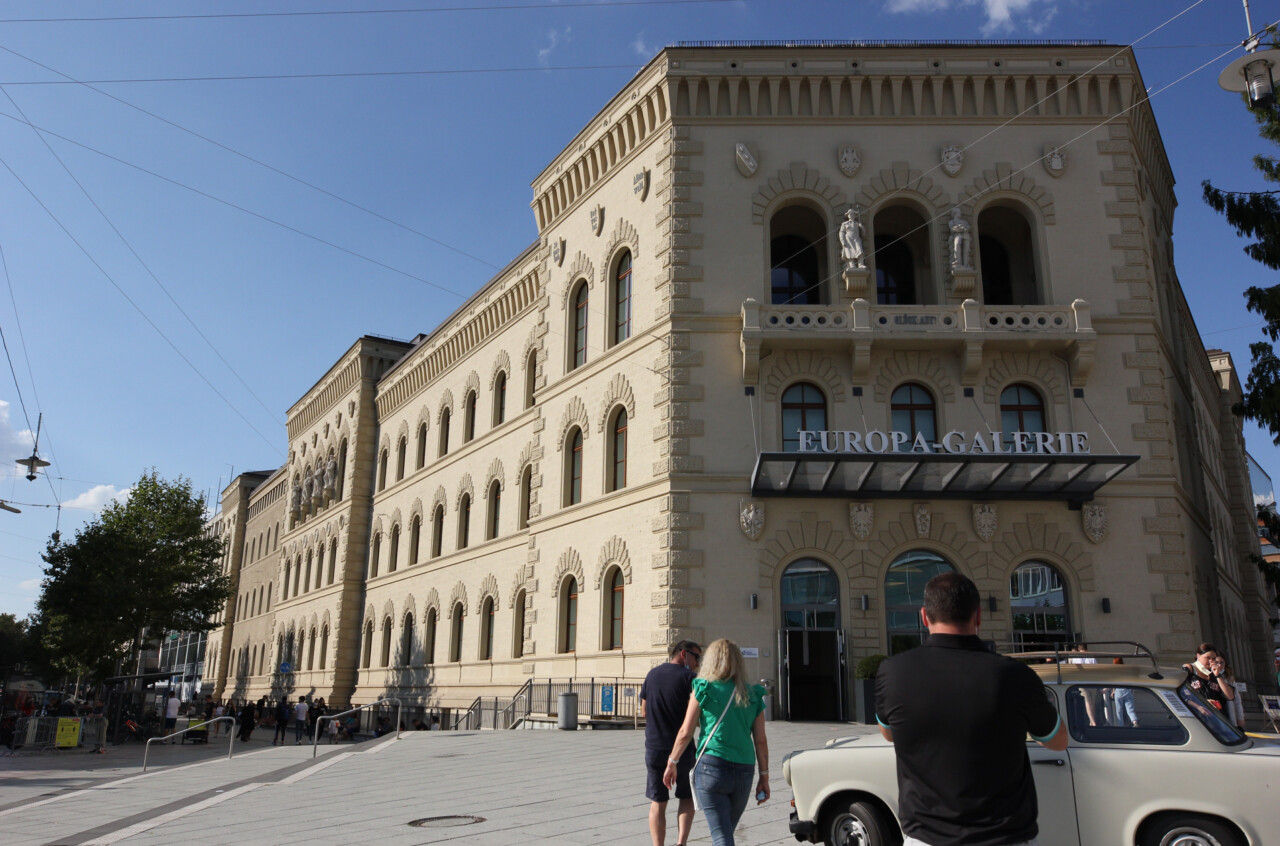}  &  
       \includegraphics[height=1.6cm,clip=true]{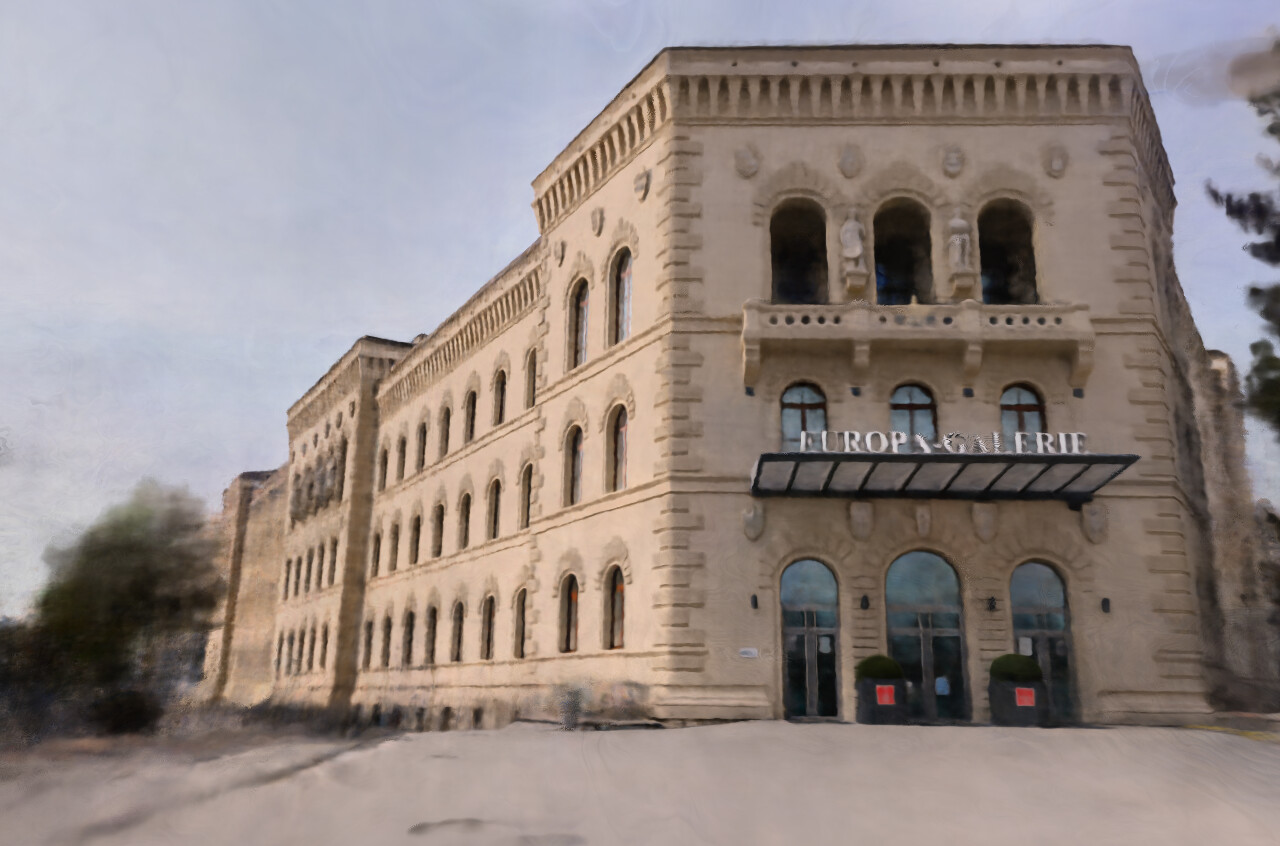}  &  
       \includegraphics[height=1.6cm,clip=true]{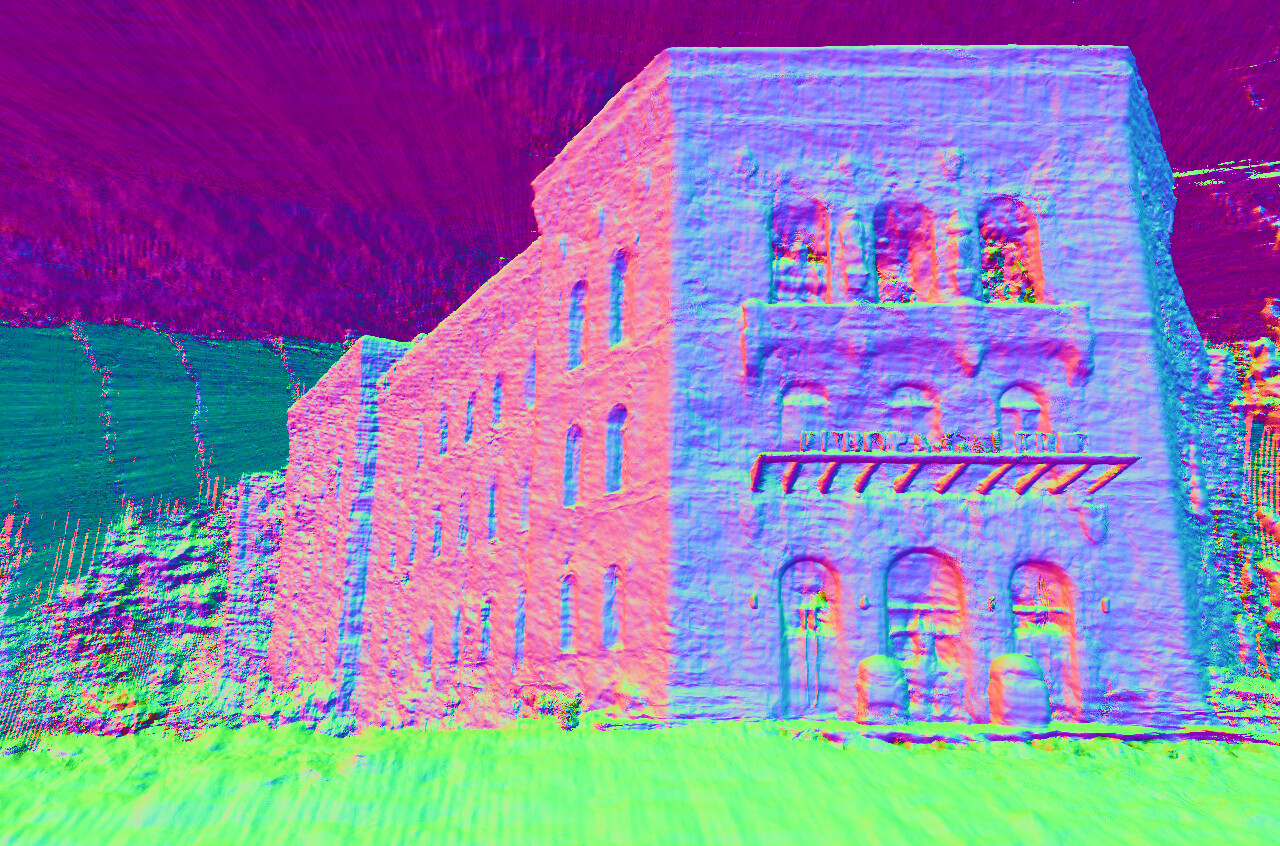}  &  
       \includegraphics[height=1.6cm,clip=true]{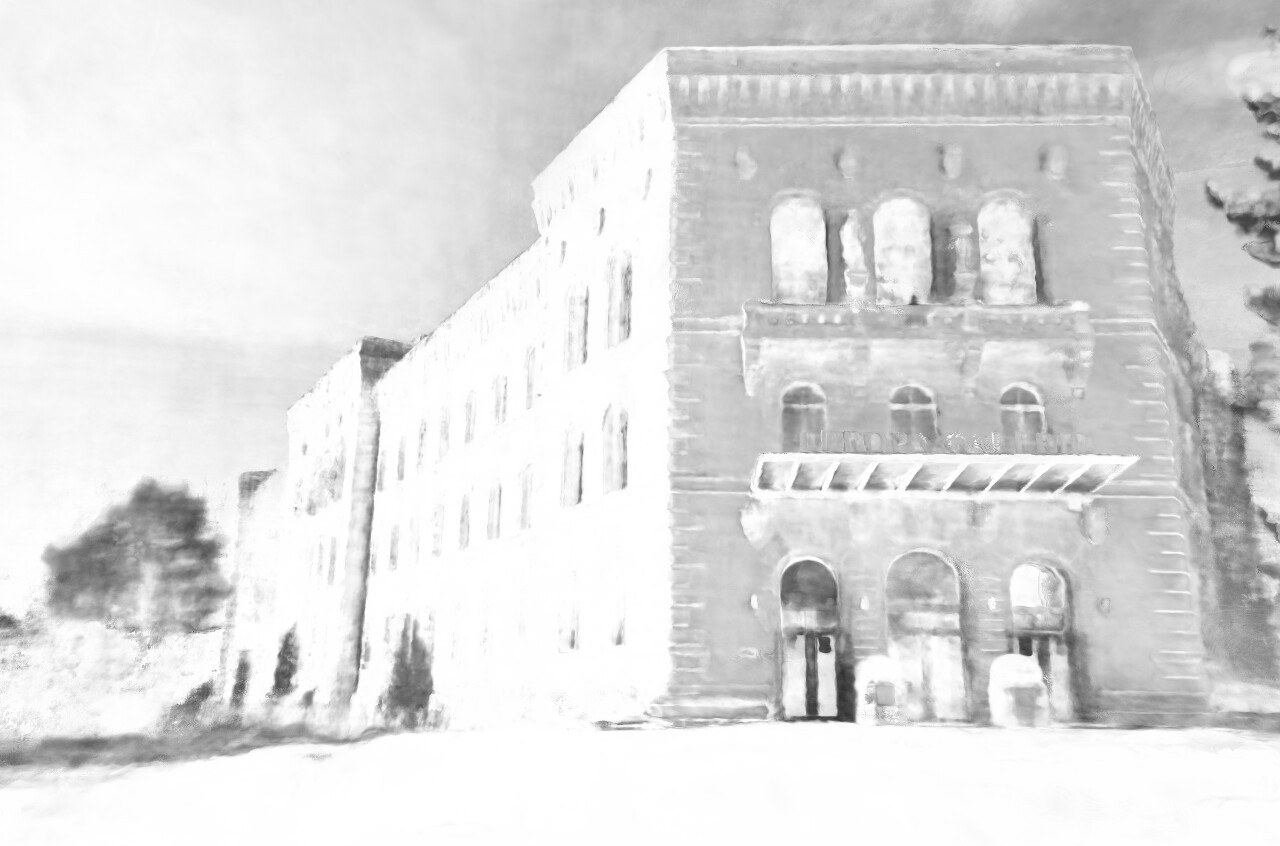}  &  
       \includegraphics[height=1.6cm,clip=true]{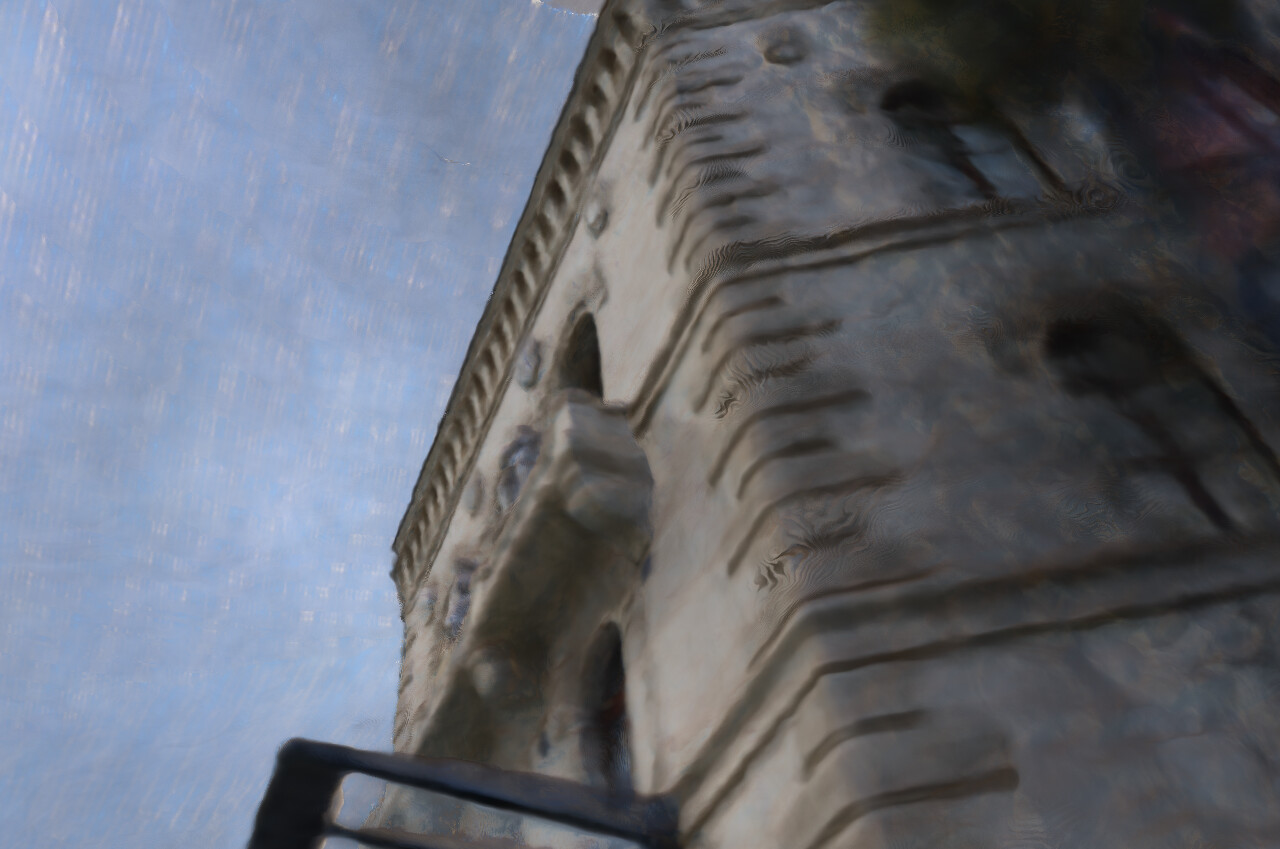}  \\
       
       \includegraphics[height=1.6cm,clip=true]{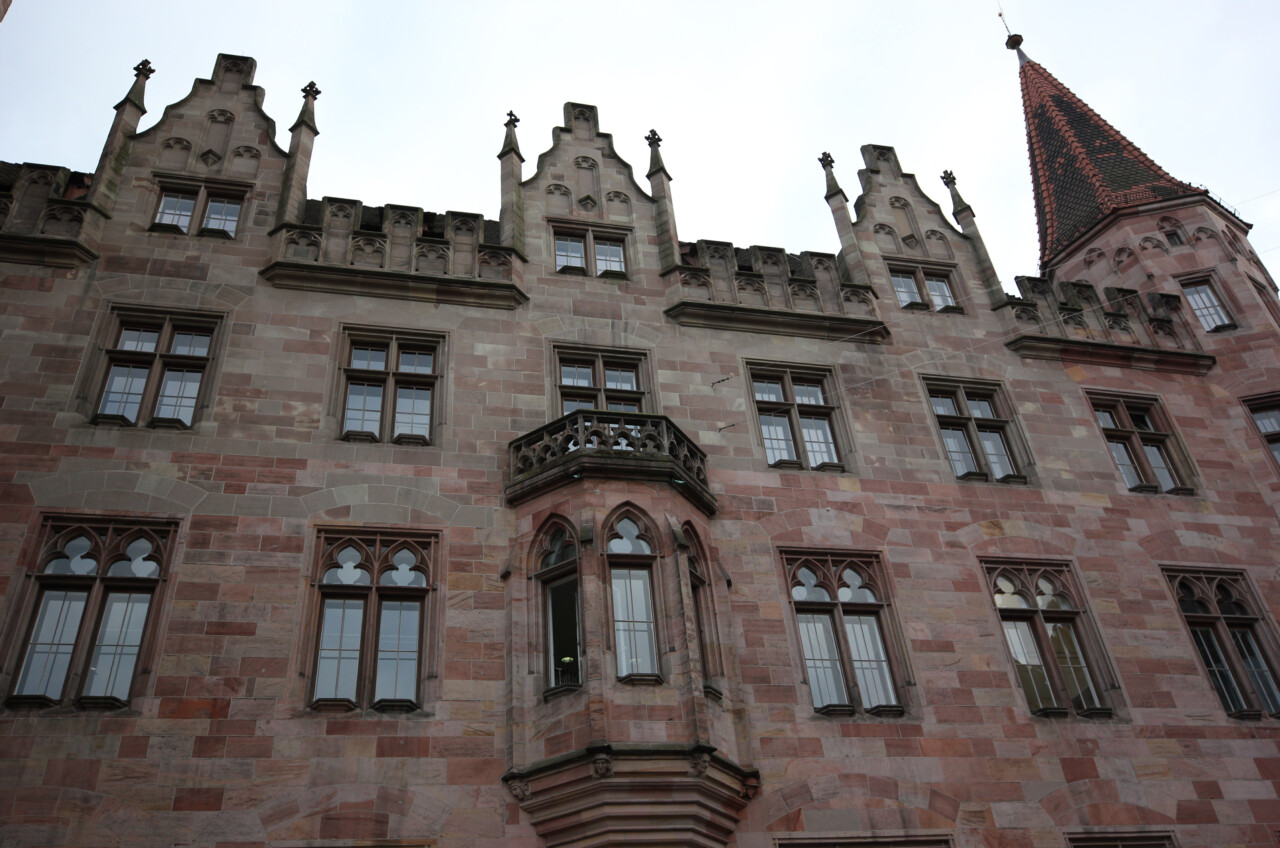}  &  
       \includegraphics[height=1.6cm,clip=true]{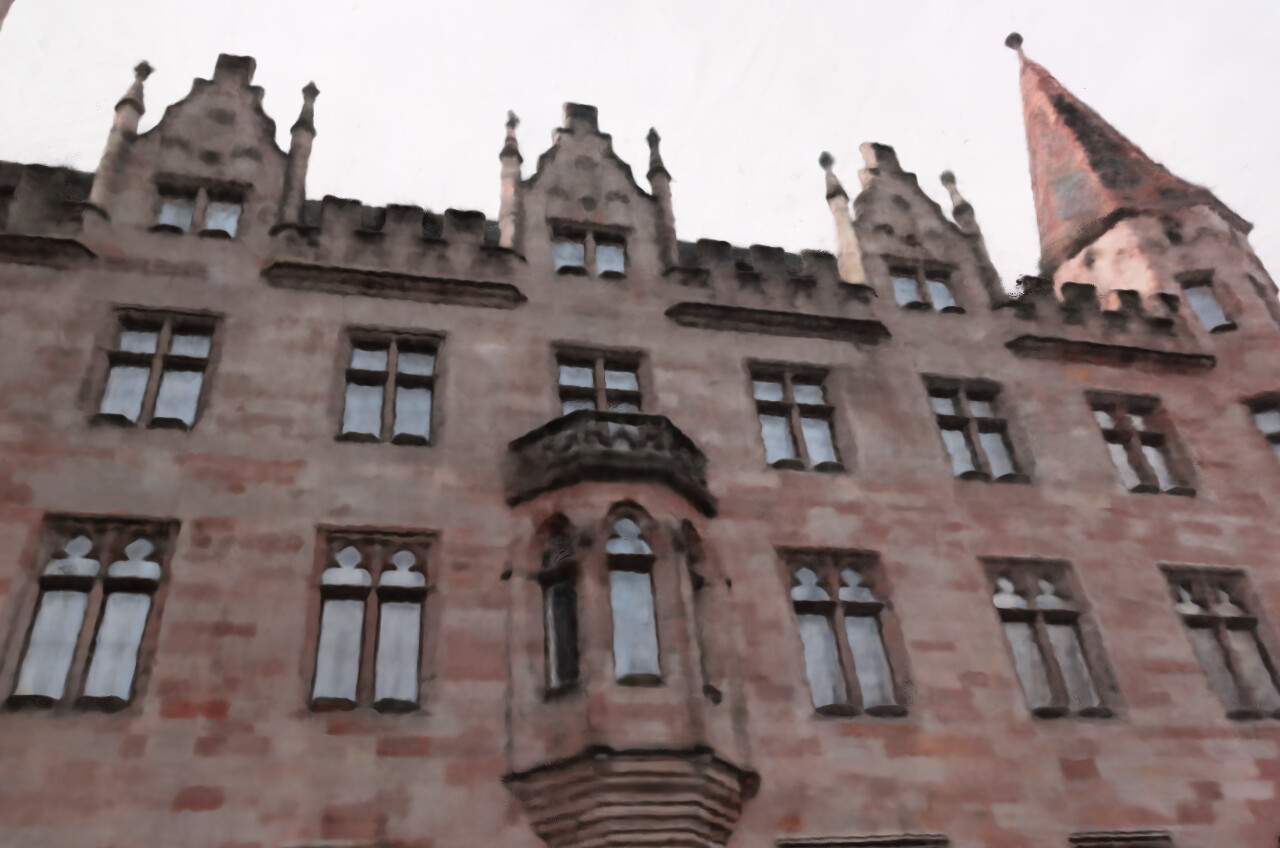}  &  
       \includegraphics[height=1.6cm,clip=true]{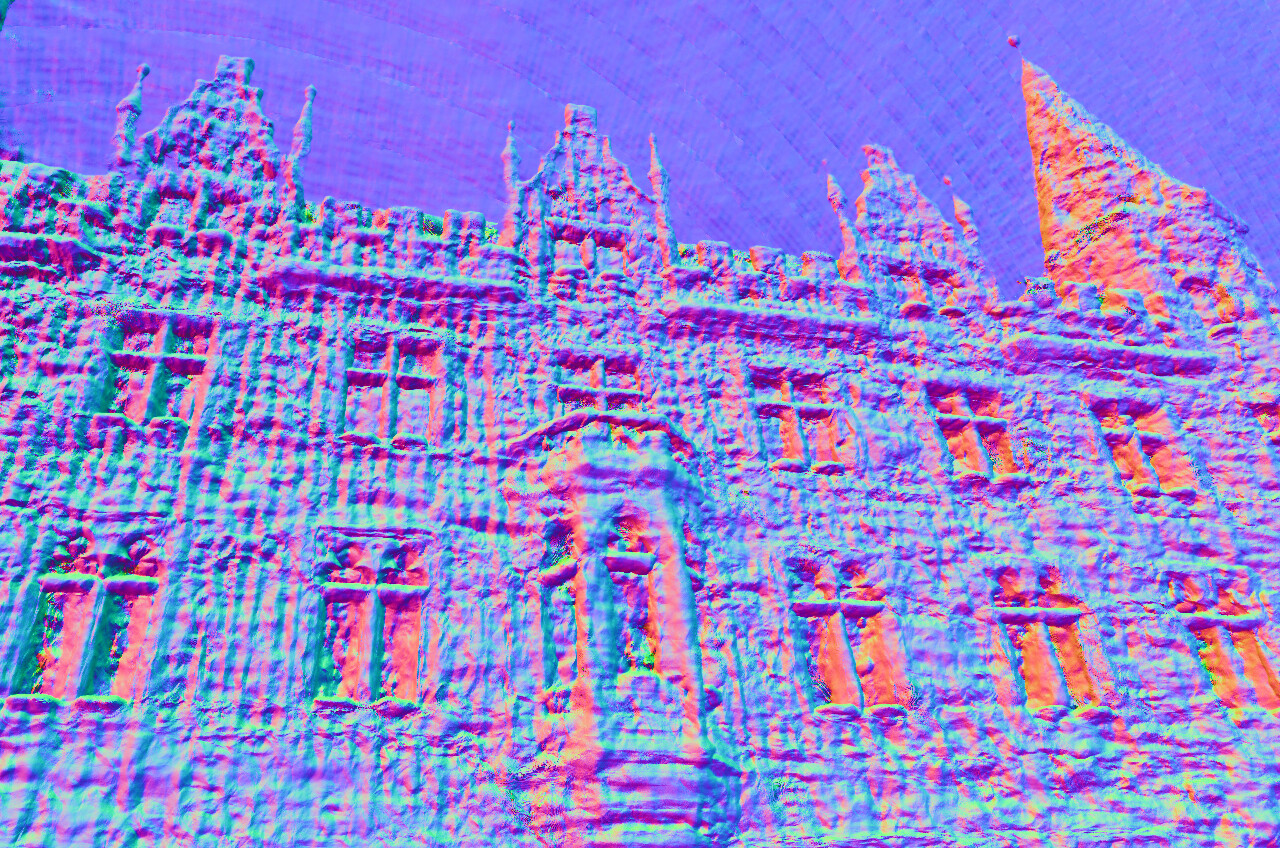}  &  
       \includegraphics[height=1.6cm,clip=true]{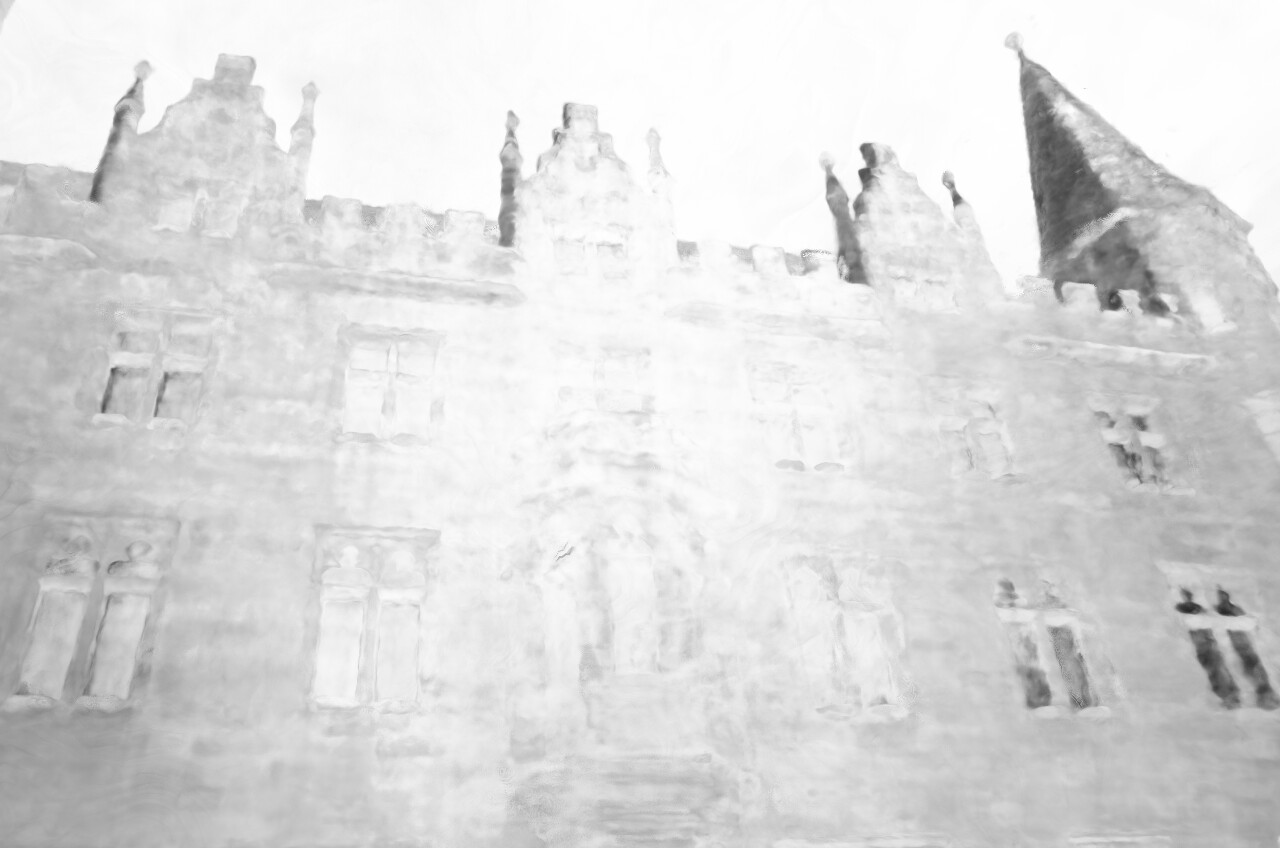}  &  
       \includegraphics[height=1.6cm,clip=true]{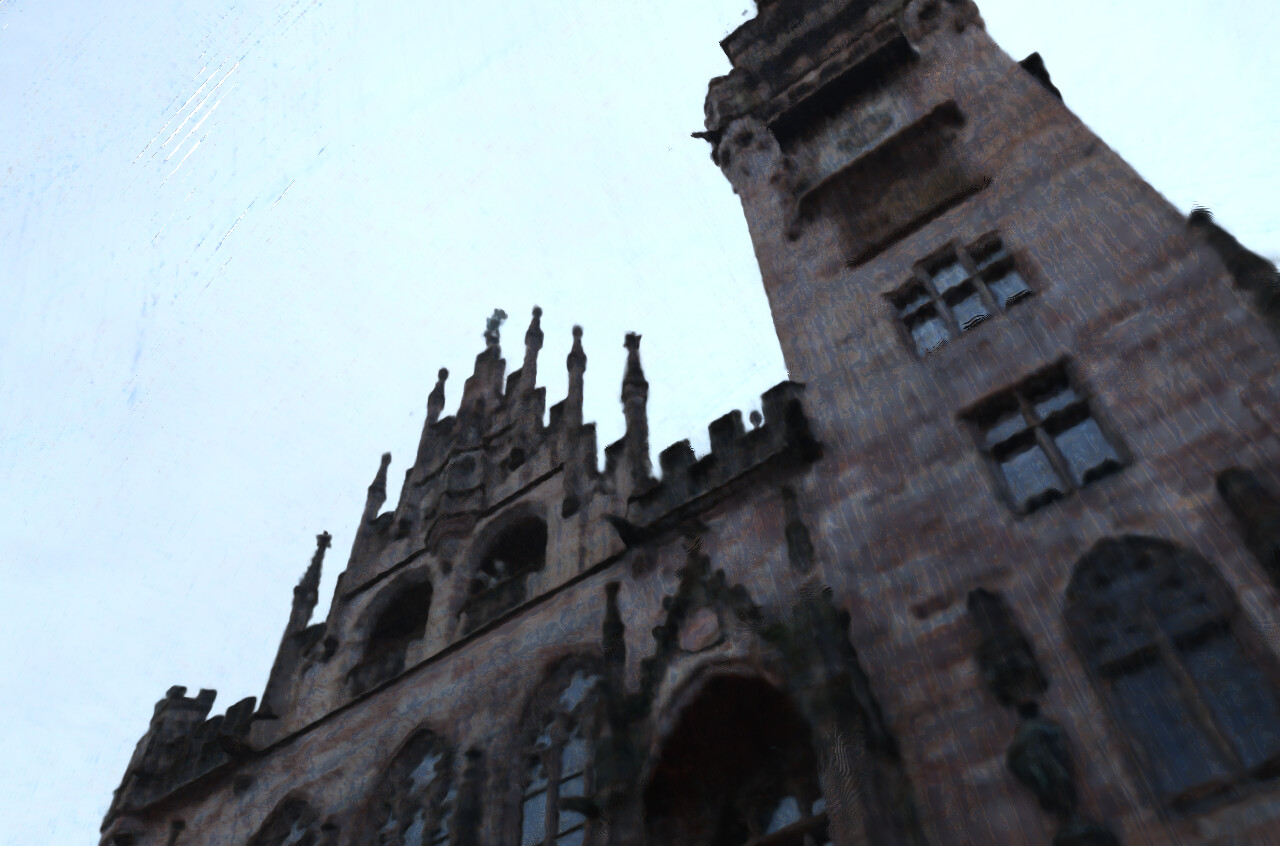}  \\
       
       \includegraphics[height=1.6cm,clip=true]{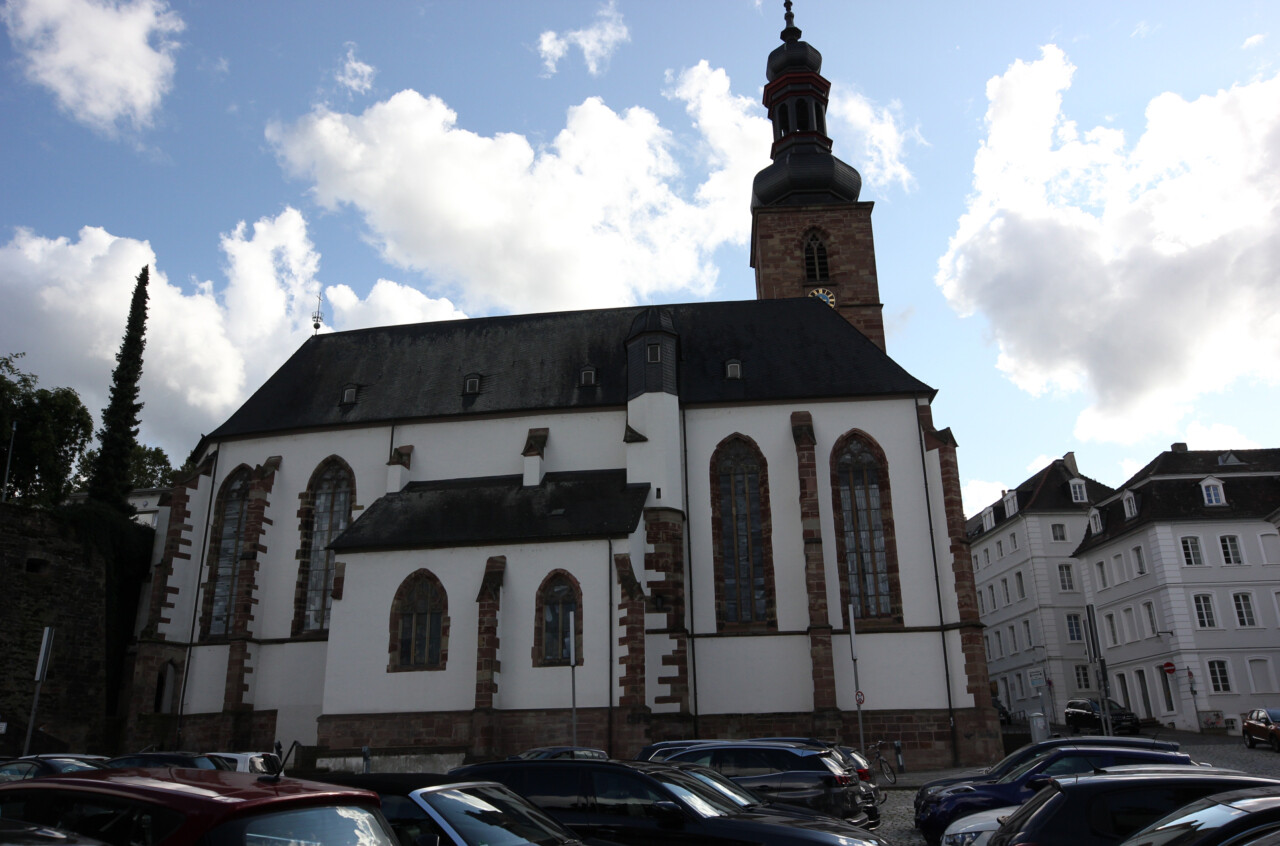}  &  
       \includegraphics[height=1.6cm,clip=true]{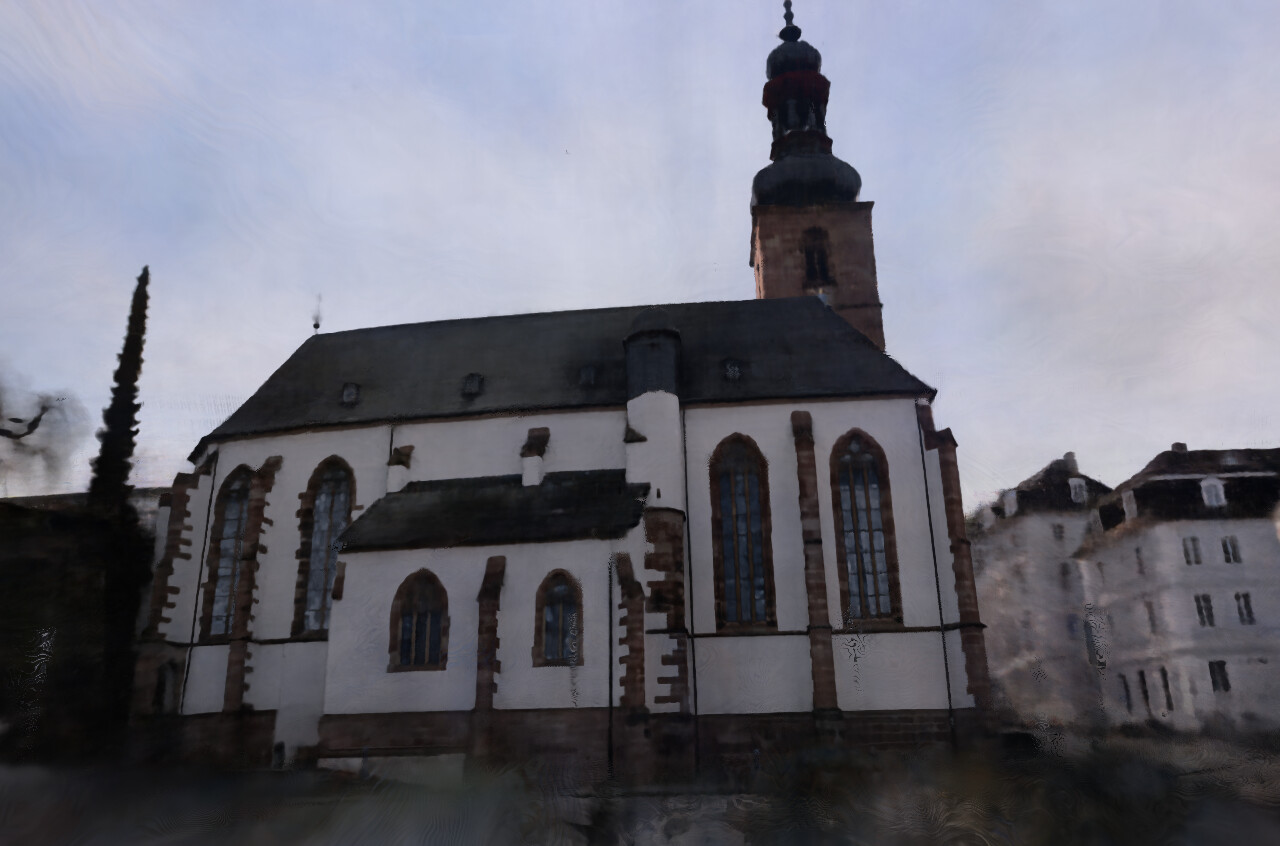}  &  
       \includegraphics[height=1.6cm,clip=true]{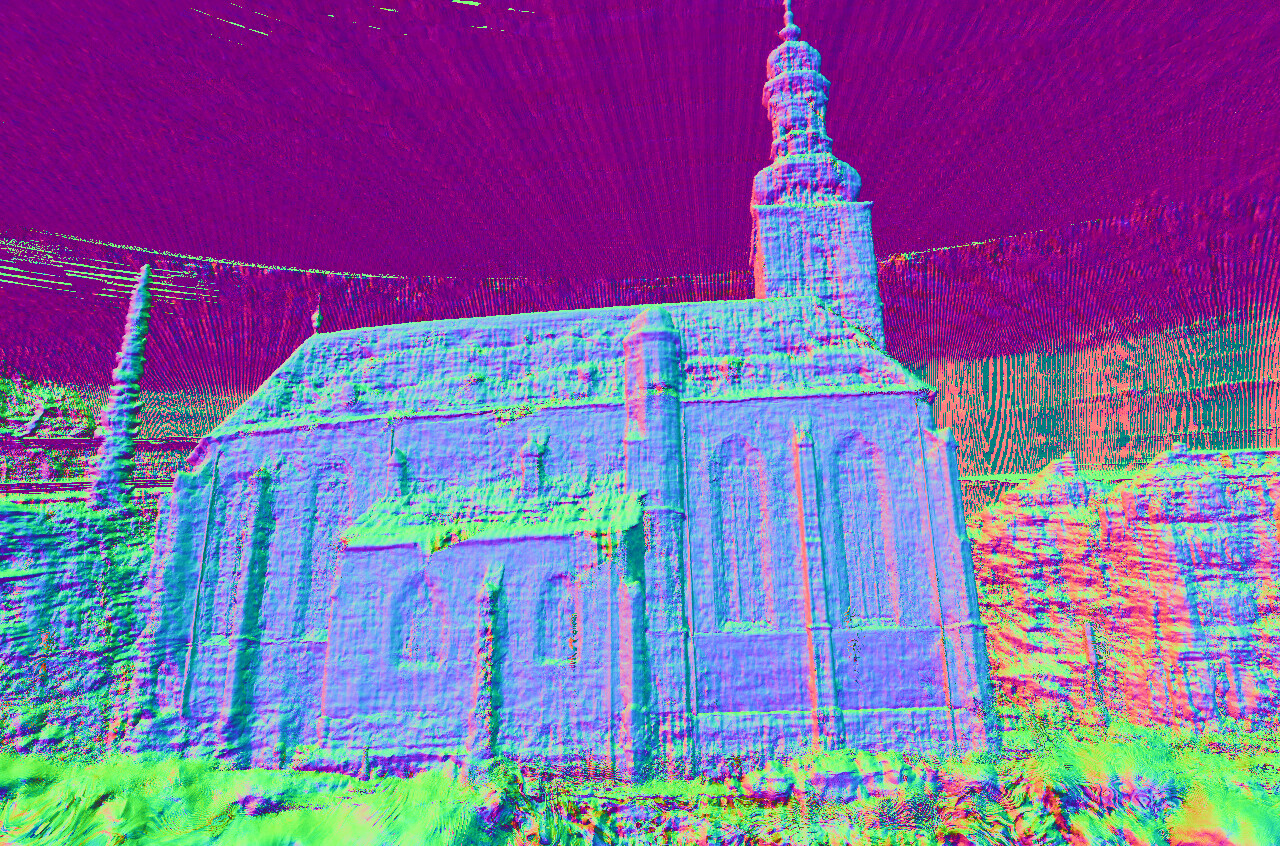}  &  
       \includegraphics[height=1.6cm,clip=true]{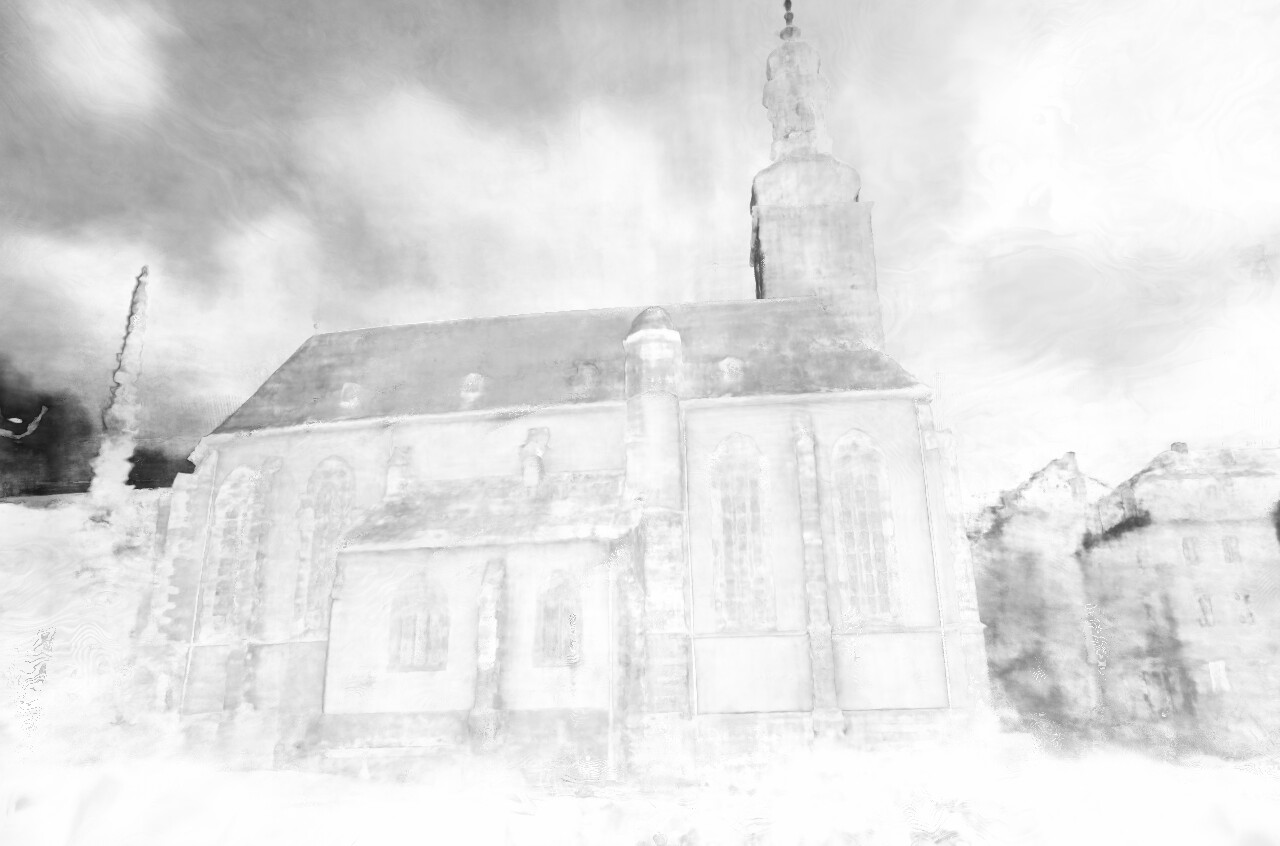}  &  
       \includegraphics[height=1.6cm,clip=true]{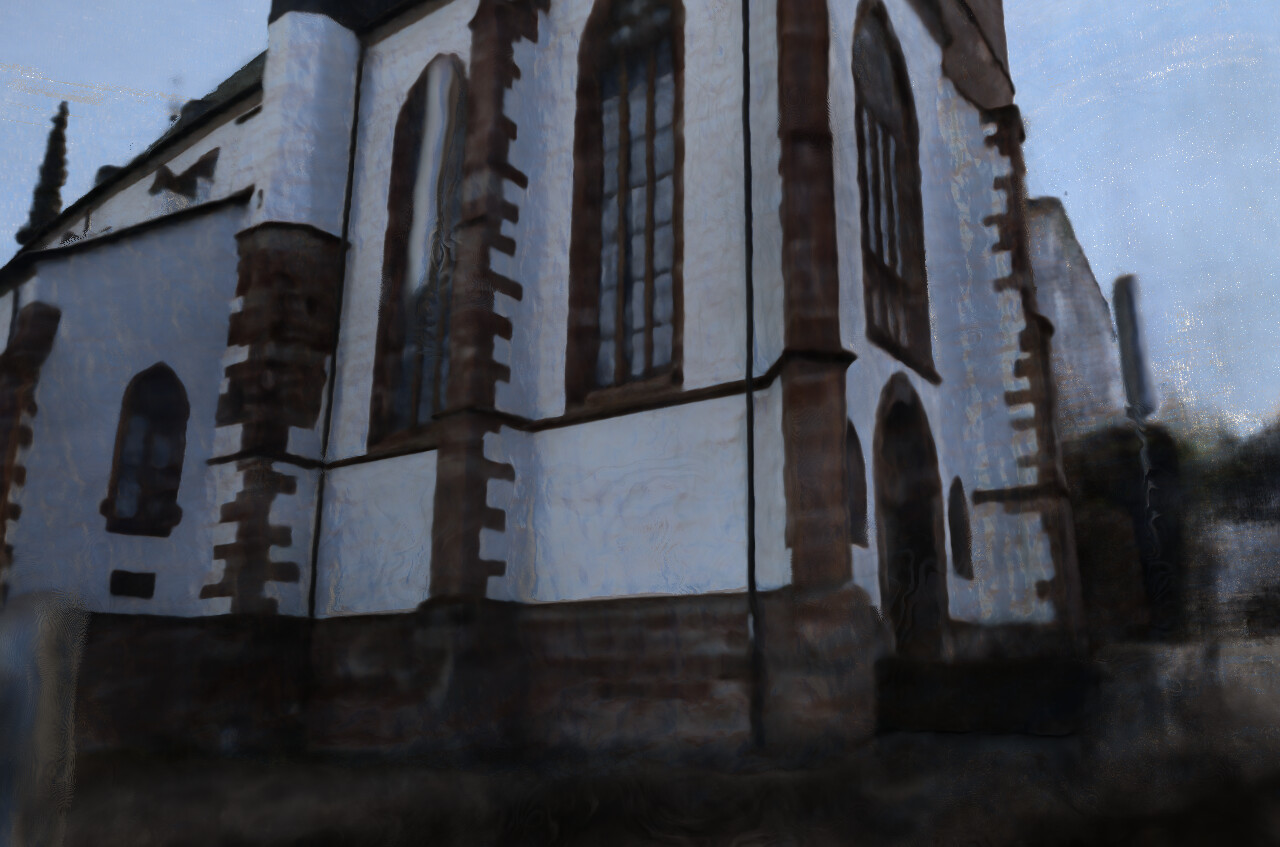}  \\
       
       \includegraphics[height=1.6cm,clip=true]{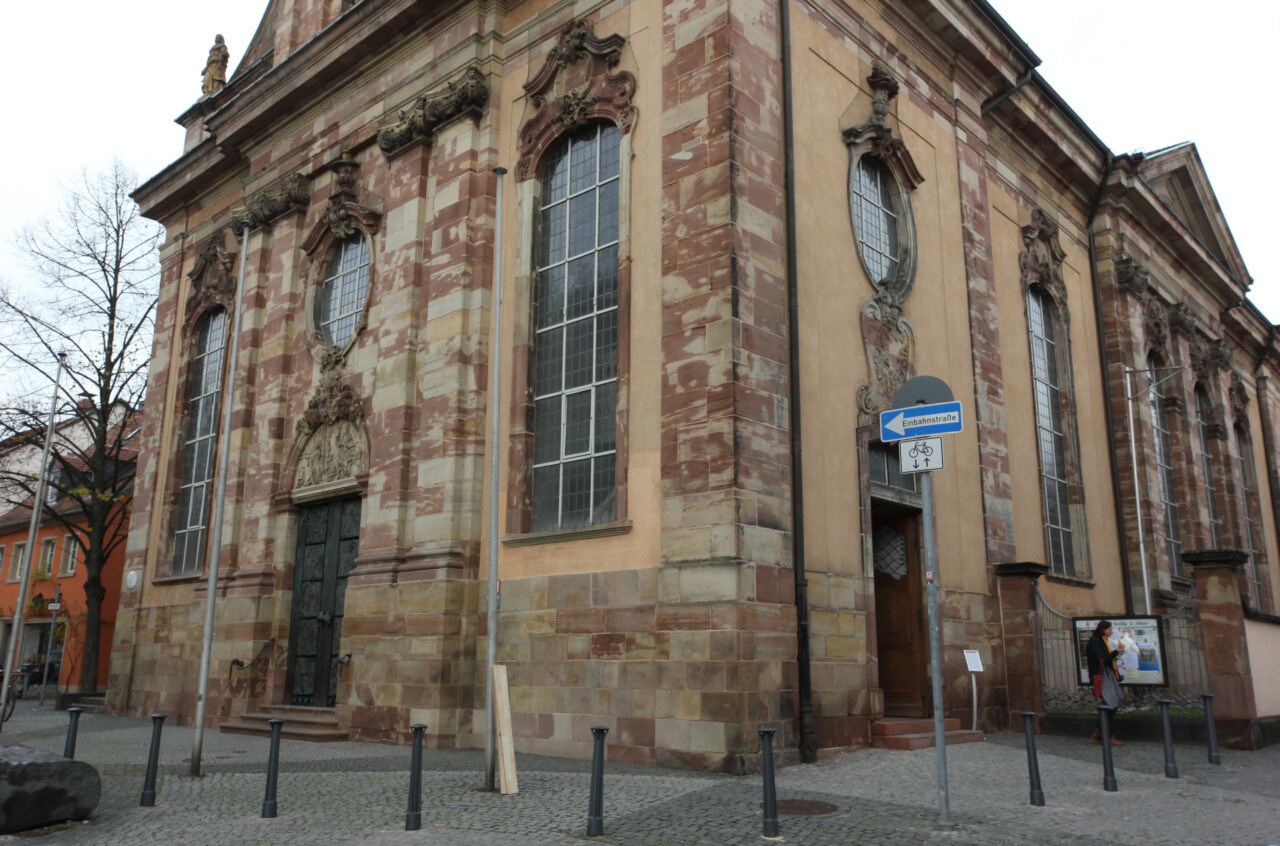}  &  
       \includegraphics[height=1.6cm,clip=true]{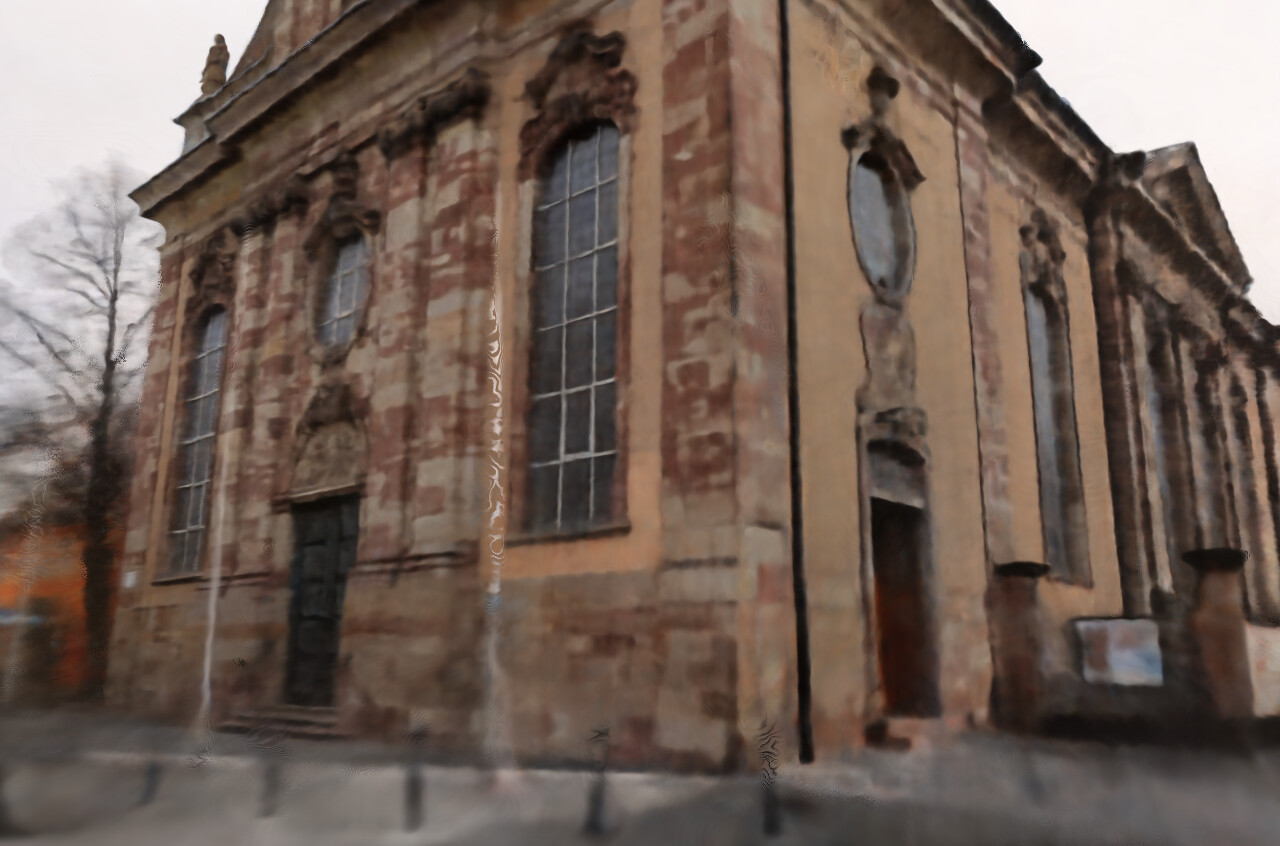}  &  
       \includegraphics[height=1.6cm,clip=true]{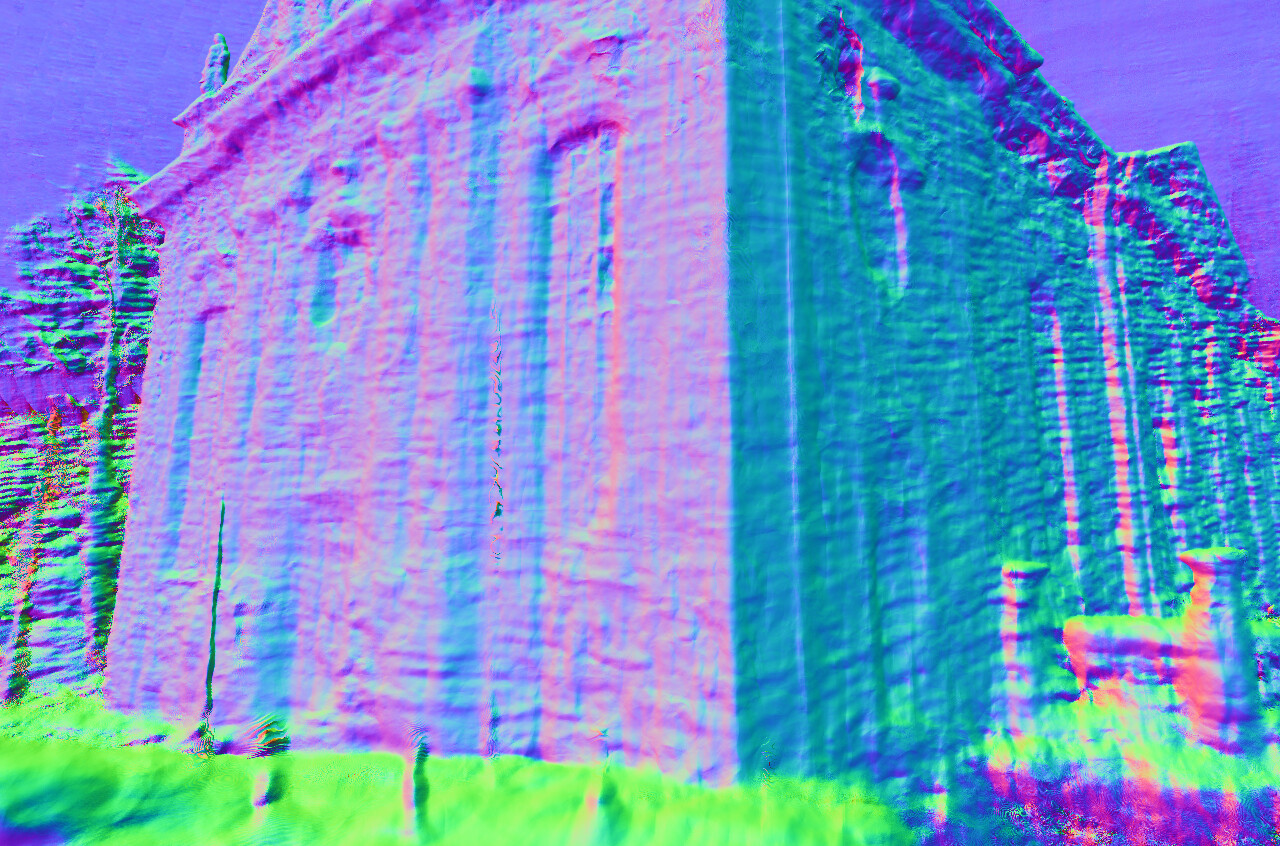}  &  
       \includegraphics[height=1.6cm,clip=true]{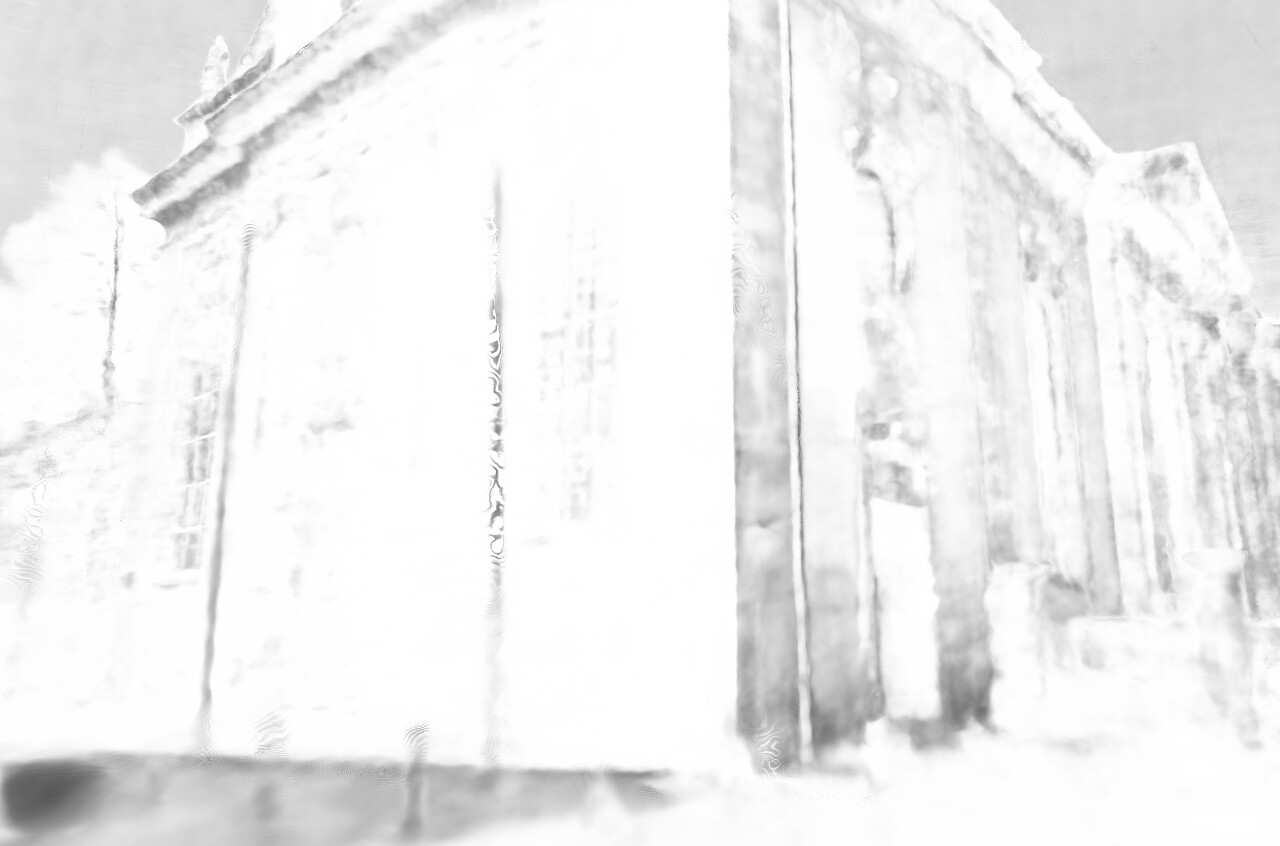}  &  
       \includegraphics[height=1.6cm,clip=true]{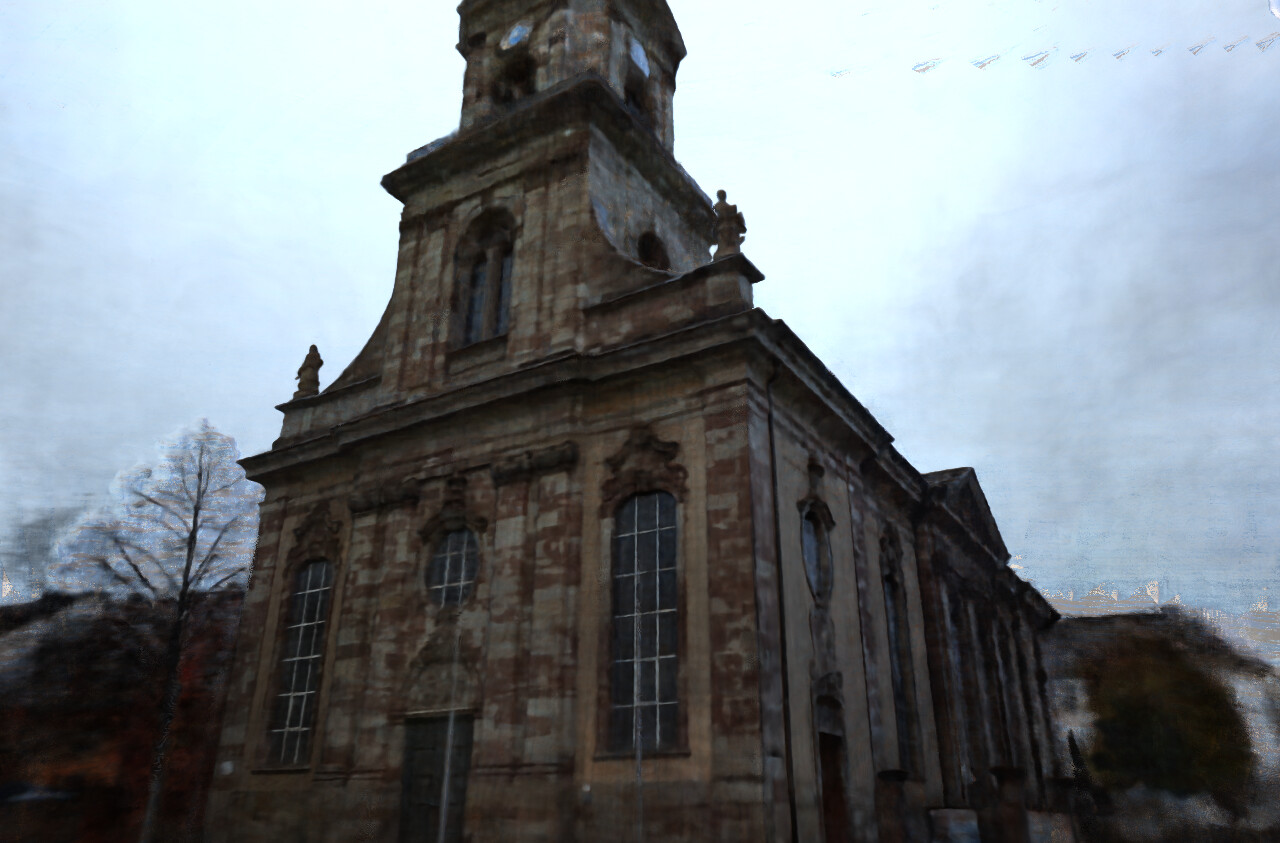}  \\
       
       \small{Training image} & \small{Diffuse Albedo} & \small{Surface Normals} & \small{Shadows} & \small{N.l. and v.} \vspace{0.1cm}
    \end{tabular}
    
    \captionof{figure}{Additional decompositions and novel lighting and viewpoint synthesis for various sites of our proposed dataset with NeRF-OSR. ``N.l. and v.'' stands for ``Novel lighting and viewpoint''. We kindly ask the readers to ignore inaccuracies in the sky as sky modelling is outside the scope of the work.
    } 
    \label{fig:moresites}
\end{figure}

Similarly to Fig.~5 of the main document, we show more reconstruction, novel view and lighting synthesis for more sites of our proposed dataset in Fig.~\ref{fig:moresites}.

\input{figures/ablativestudycol}

\input{figures/additionalvis}

%% file: figures/dataset_tbl.tex
\begin{table}[h]
    \centering
    \begin{tabular}{c|c|c}
                & Sessions & Views\\
         \hline\hline
         Site 1 & 18 & 373\\
         Site 2 & 17 & 423\\
         Site 3 & 16 & 372\\
         Site 4 & 11 & 401 \\
         \hline
         \multicolumn{3}{c}{}
    \end{tabular}
    \begin{tabular}{c|c|c}
                & Sessions & Views\\
         \hline\hline
         Site 5 & 13 & 493 \\
         Site 6 & 12 & 379\\
         Site 7 & 11 & 468\\
         Site 8 & 12 & 331\\
         \hline
         Total & 110 & 3240
    \end{tabular}
    \caption{Statistics of our new benchmark dataset. 
    The dataset currently contains eight sites recorded in 110 different sessions, each with a $360^\circ$ environment map captured by LG R105. 
    The total number of views captured by a DSLR camera Canon EOS 550D is 3240. 
    } 
    \label{tbl:dataset_summary} 
\end{table}

%% file: figures/ablativestudycol.tex
\begin{figure*}
    \centering
    \includegraphics[width=1.0\linewidth]{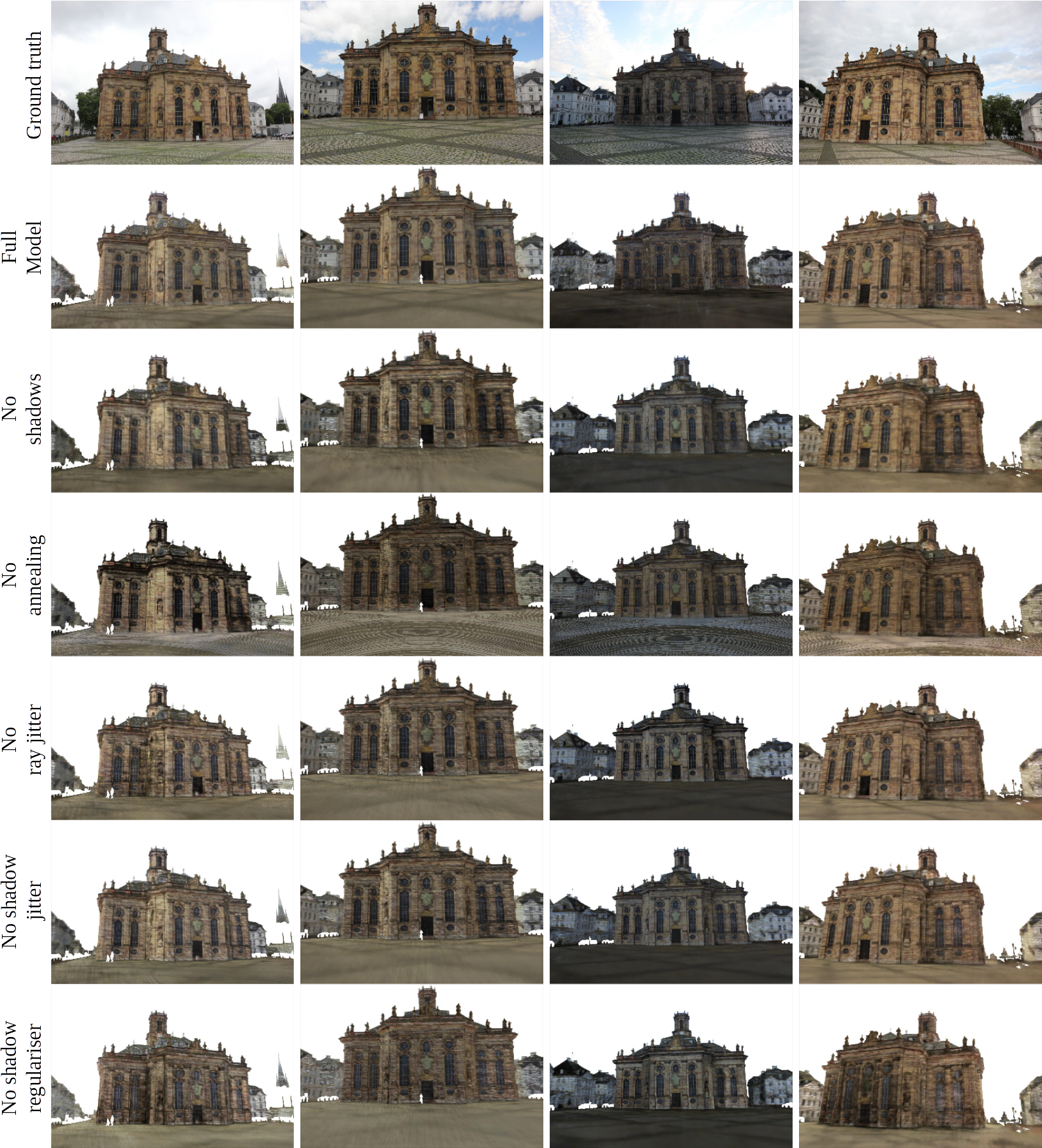}
\caption{
The impact of the various design choices in NeRF-OSR (Site 1). 
The four columns show the view-lighting combinations used in the quantitative evaluation against the ground truth (Tab.~1 of the main manuscript). 
The best result is obtained using the full model (the second row from the top). Best viewed with zoom. 
}
    \label{fig:ablationCol}
\end{figure*}

%% file: figures/additionalvis.tex
\begin{figure*}
    \centering
    
    \includegraphics[width=\textwidth]{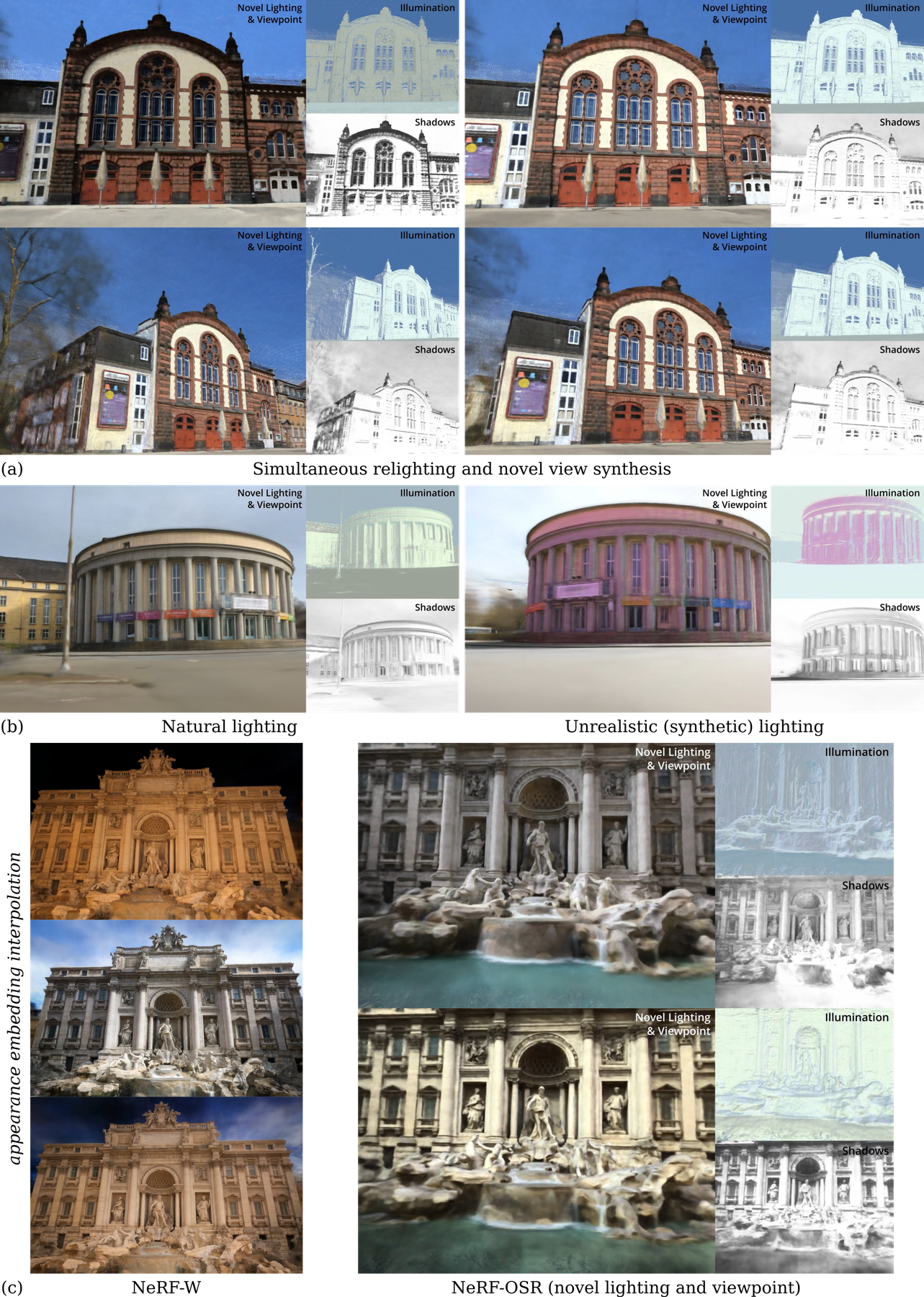}
    \captionof{figure}{Additional visualisations for various experiments. 
    (a): Relighting and novel view synthesis of Site 3; (b): Relighting of Site 2 using natural and unrealistic light sources (illuminations); (c): Qualitative comparisons to NeRF-W \cite{Brualla20nerfw}. 
    For the corresponding full videos, see our supplementary video. 
    } 
    \label{fig:additionalvis}
\end{figure*}

%% file: eccv2022submission.bbl
\begin{thebibliography}{10}
\providecommand{\url}[1]{\texttt{#1}}
\providecommand{\urlprefix}{URL }
\providecommand{\doi}[1]{https://doi.org/#1}

\bibitem{Barron15}
{Barron}, J.T., {Malik}, J.: Shape, illumination, and reflectance from shading.
  IEEE Transactions on Pattern Analysis and Machine Intelligence (TPAMI)
  \textbf{37}(8),  1670--1687 (2015)

\bibitem{basri2003lambertian}
Basri, R., Jacobs, D.W.: Lambertian reflectance and linear subspaces. IEEE
  TPAMI  \textbf{25}(2),  218--233 (2003)

\bibitem{Boss20NeRD}
Boss, M., Braun, R., Jampani, V., Barron, J.T., Liu, C., Lensch, H.P.: Nerd:
  Neural reflectance decomposition from image collections. In: International
  Conference on Computer Vision (ICCV) (2021)

\bibitem{chen2022hallucinated}
Chen, X., Zhang, Q., Li, X., Chen, Y., Feng, Y., Wang, X., Wang, J.:
  Hallucinated neural radiance fields in the wild. In: Proceedings of the
  IEEE/CVF Conference on Computer Vision and Pattern Recognition. pp.
  12943--12952 (2022)

\bibitem{Duchene15}
Duch\^{e}ne, S., Riant, C., Chaurasia, G., Moreno, J.L., Laffont, P.Y., Popov,
  S., Bousseau, A., Drettakis, G.: Multiview intrinsic images of outdoors
  scenes with an application to relighting. ACM Trans. Graph.  \textbf{34}(5)
  (2015)

\bibitem{Finlayson15}
Finlayson, G.D., Mackiewicz, M., Hurlbert, A.: Color correction using
  root-polynomial regression. IEEE Transactions on Image Processing
  \textbf{24}(5),  1460--1470 (2015)

\bibitem{Garon19}
Garon, M., Sunkavalli, K., Hadap, S., Carr, N., Lalonde, J.F.: Fast
  spatially-varying indoor lighting estimation. In: Computer Vision and Pattern
  Recognition (CVPR) (2019)

\bibitem{Guo19}
Guo, K., Lincoln, P., Davidson, P., Busch, J., Yu, X., Whalen, M., Harvey, G.,
  Orts-Escolano, S., Pandey, R., Dourgarian, J., Tang, D., Tkach, A., Kowdle,
  A., Cooper, E., Dou, M., Fanello, S., Fyffe, G., Rhemann, C., Taylor, J.,
  Debevec, P., Izadi, S.: The relightables: Volumetric performance capture of
  humans with realistic relighting. ACM Trans. Graph.  \textbf{38}(6) (2019)

\bibitem{Jin2020}
Jin, Y., Mishkin, D., Mishchuk, A., Matas, J., Fua, P., Yi, K.M., Trulls, E.:
  {Image Matching across Wide Baselines: From Paper to Practice}. International
  Journal of Computer Vision (IJCV)  (2020)

\bibitem{Karsch11}
Karsch, K., Hedau, V., Forsyth, D., Hoiem, D.: Rendering synthetic objects into
  legacy photographs. ACM Trans. Graph.  \textbf{30}(6) (2011)

\bibitem{Laffont12}
Laffont, P.Y., Bousseau, A., Paris, S., Durand, F., Drettakis, G.: Coherent
  intrinsic images from photo collections. ACM Trans. Graph.  \textbf{31}(6)
  (2012)

\bibitem{Lalonde09}
Lalonde, J.F., Efros, A.A., Narasimhan, S.G.: Webcam clip art: Appearance and
  illuminant transfer from time-lapse sequences. ACM Trans. Graph.
  \textbf{28}(5),  1–10 (Dec 2009)

\bibitem{Li12}
Li, Y., Snavely, N., Huttenlocher, D., Fua, P.: Worldwide pose estimation using
  3d point clouds. In: European Conference on Computer Vision (ECCV). pp.
  15--29 (2012)

\bibitem{MegaDepthLi18}
Li, Z., Snavely, N.: Megadepth: Learning single-view depth prediction from
  internet photos. In: Computer Vision and Pattern Recognition (CVPR) (2018)

\bibitem{Li20crowdsampling}
Li, Z., Xian, W., Davis, A., Snavely, N.: Crowdsampling the plenoptic function.
  In: European Conference on Computer Vision (ECCV) (2020)

\bibitem{marchingcubes}
Lorensen, W.E., Cline, H.E.: Marching cubes: A high resolution 3d surface
  construction algorithm. ACM siggraph computer graphics  \textbf{21}(4),
  163--169 (1987)

\bibitem{Luan17}
{Luan}, F., {Paris}, S., {Shechtman}, E., {Bala}, K.: Deep photo style
  transfer. In: Computer Vision and Pattern Recognition (CVPR). pp. 6997--7005
  (2017)

\bibitem{mallikarjun2021photoapp}
Mallikarjun, B.R., Tewari, A., Dib, A., Weyrich, T., Bickel, B., Seidel, H.P.,
  Pfister, H., Matusik, W., Chevallier, L., Elgharib, M., et~al.: Photoapp:
  Photorealistic appearance editing of head portraits. ACM Transactions on
  Graphics  \textbf{40}(4) (2021)

\bibitem{Brualla20nerfw}
Martin-Brualla, R., Radwan, N., Sajjadi, M.S.M., Barron, J.T., Dosovitskiy, A.,
  Duckworth, D.: {NeRF in the Wild: Neural Radiance Fields for Unconstrained
  Photo Collections}. In: Computer Vision and Pattern Recognition (CVPR) (2021)

\bibitem{Meka2018}
Meka, A., Maximov, M., Zollhoefer, M., Chatterjee, A., Seidel, H.P., Richardt,
  C., Theobalt, C.: Lime: Live intrinsic material estimation. In: Computer
  Vision and Pattern Recognition ({CVPR}) (2018)

\bibitem{Meka20}
Meka, A., Pandey, R., Haene, C., Orts-Escolano, S., Barnum, P., Davidson, P.,
  Erickson, D., Zhang, Y., Taylor, J., Bouaziz, S., Legendre, C., Ma, W.C.,
  Overbeck, R., Beeler, T., Debevec, P., Izadi, S., Theobalt, C., Rhemann, C.,
  Fanello, S.: Deep relightable textures - volumetric performance capture with
  neural rendering. In: ACM Transactions on Graphics (Proceedings SIGGRAPH
  Asia). vol.~39 (2020)

\bibitem{Meshry19}
{Meshry}, M., {Goldman}, D.B., {Khamis}, S., {Hoppe}, H., {Pandey}, R.,
  {Snavely}, N., {Martin-Brualla}, R.: Neural rerendering in the wild. In:
  Computer Vision and Pattern Recognition (CVPR). pp. 6871--6880 (2019)

\bibitem{Mildenhall20nerf}
Mildenhall, B., Srinivasan, P.P., Tancik, M., Barron, J.T., Ramamoorthi, R.,
  Ng, R.: Nerf: Representing scenes as neural radiance fields for view
  synthesis. In: European Conference on Computer Vision (ECCV) (2020)

\bibitem{Nam19}
Nam, S., Ma, C., Chai, M., Brendel, W., Xu, N., Kim, S.: End-to-end time-lapse
  video synthesis from a single outdoor image. Computer Vision and Pattern
  Recognition (CVPR) pp. 1409--1418 (2019)

\bibitem{Oechsle2021ICCV}
Oechsle, M., Peng, S., Geiger, A.: Unisurf: Unifying neural implicit surfaces
  and radiance fields for multi-view reconstruction. In: International
  Conference on Computer Vision (ICCV) (2021)

\bibitem{park2020nerfies}
Park, K., Sinha, U., Barron, J.T., Bouaziz, S., Goldman, D.B., Seitz, S.M.,
  Martin-Brualla, R.: Nerfies: Deformable neural radiance fields. International
  Conference on Computer Vision (ICCV)  (2021)

\bibitem{Philip19}
Philip, J., Gharbi, M., Zhou, T., Efros, A.A., Drettakis, G.: Multi-view
  relighting using a geometry-aware network. ACM Trans. Graph.  \textbf{38}(4)
  (2019)

\bibitem{Schonberger16}
Sch{\"o}nberger, J.L., Zheng, E., Frahm, J.M., Pollefeys, M.: Pixelwise view
  selection for unstructured multi-view stereo. In: European Conference on
  Computer Vision (ECCV). pp. 501--518 (2016)

\bibitem{Schonberger16colmap}
Schönberger, J.L., Frahm, J.M.: Structure-from-motion revisited. In: Computer
  Vision and Pattern Recognition (CVPR). pp. 4104--4113 (2016)

\bibitem{Sengupta19ICCV}
Sengupta, S., Gu, J., Kim, K., Liu, G., Jacobs, D.W., Kautz, J.: Neural inverse
  rendering of an indoor scene from a single image. In: International
  Conference on Computer Vision (ICCV) (2019)

\bibitem{Shih13}
Shih, Y., Paris, S., Durand, F., Freeman, W.T.: Data-driven hallucination of
  different times of day from a single outdoor photo. ACM Trans. Graph.
  \textbf{32}(6) (2013)

\bibitem{Snavely06}
Snavely, N., Seitz, S.M., Szeliski, R.: Photo tourism: Exploring photo
  collections in 3d. ACM Trans. Graph.  \textbf{25}(3),  835–846 (2006)

\bibitem{Srinivasan21NeRV}
Srinivasan, P.P., Deng, B., Zhang, X., Tancik, M., Mildenhall, B., Barron,
  J.T.: Nerv: Neural reflectance and visibility fields for relighting and view
  synthesis. In: Computer Vision and Pattern Recognition (CVPR) (2021)

\bibitem{Sun19}
Sun, T., Barron, J.T., Tsai, Y.T., Xu, Z., Yu, X., Fyffe, G., Rhemann, C.,
  Busch, J., Debevec, P., Ramamoorthi, R.: Single image portrait relighting.
  ACM Trans. Graph.  \textbf{38}(4) (Jul 2019)

\bibitem{sun2021nelf}
Sun, T., Lin, K.E., Bi, S., Xu, Z., Ramamoorthi, R.: Nelf: Neural
  light-transport field for portrait view synthesis and relighting. In:
  Eurographics Symposium on Rendering (2021)

\bibitem{Sunkavalli07}
Sunkavalli, K., Matusik, W., Pfister, H., Rusinkiewicz, S.: Factored time-lapse
  video. ACM Trans. Graph.  \textbf{26}(3) (2007)

\bibitem{tancik2022blocknerf}
Tancik, M., Casser, V., Yan, X., Pradhan, S., Mildenhall, B., Srinivasan, P.,
  Barron, J.T., Kretzschmar, H.: {Block-NeRF}: Scalable large scene neural view
  synthesis. In: Proceedings of the IEEE/CVF Conference on Computer Vision and
  Pattern Recognition. pp. 8248--8258 (2022)

\bibitem{Tao20}
Tao, A., Sapra, K., Catanzaro, B.: Hierarchical multi-scale attention for
  semantic segmentation. arXiv preprint arXiv:2005.10821  (2020)

\bibitem{2021arXiv211105849T}
{Tewari}, A., {Thies}, J., {Mildenhall}, B., {Srinivasan}, P., {Tretschk}, E.,
  {Wang}, Y., {Lassner}, C., {Sitzmann}, V., {Martin-Brualla}, R., {Lombardi},
  S., {Simon}, T., {Theobalt}, C., {Niessner}, M., {Barron}, J.T., {Wetzstein},
  G., {Zollhoefer}, M., {Golyanik}, V.: {Advances in Neural Rendering}. arXiv
  e-prints  (2021)

\bibitem{tretschk2021nonrigid}
Tretschk, E., Tewari, A., Golyanik, V., Zollh\"{o}fer, M., Lassner, C.,
  Theobalt, C.: Non-rigid neural radiance fields: Reconstruction and novel view
  synthesis of a dynamic scene from monocular video. In: International
  Conference on Computer Vision ({ICCV}) (2021)

\bibitem{scikitimage}
Van~der Walt, S., Sch{\"o}nberger, J.L., Nunez-Iglesias, J., Boulogne, F.,
  Warner, J.D., Yager, N., Gouillart, E., Yu, T.: scikit-image: image
  processing in python. PeerJ  \textbf{2}, ~e453 (2014)

\bibitem{wang2021neus}
Wang, P., Liu, L., Liu, Y., Theobalt, C., Komura, T., Wang, W.: Neus: Learning
  neural implicit surfaces by volume rendering for multi-view reconstruction.
  Neural Information Processing Systems (NeurIPS)  (2021)

\bibitem{shadowmap}
Williams, L.: Casting curved shadows on curved surfaces. In: Proceedings of the
  5th annual conference on Computer graphics and interactive techniques. pp.
  270--274 (1978)

\bibitem{Xing13}
Xing, G., Zhou, X., Peng, Q., Liu, Y., Qin, X.: {Lighting Simulation of
  Augmented Outdoor Scene Based on a Legacy Photograph}. Computer Graphics
  Forum  (2013)

\bibitem{Xu18deep}
Xu, Z., Sunkavalli, K., Hadap, S., Ramamoorthi, R.: Deep image-based relighting
  from optimal sparse samples. ACM Transactions on Graphics  \textbf{37}(4),
  ~126 (2018)

\bibitem{Yu20}
Yu, Y., Meka, A., Elgharib, M., Seidel, H.P., Theobalt, C., Smith, W.:
  Self-supervised outdoor scene relighting. In: European Conference on Computer
  Vision (ECCV) (2020)

\bibitem{Yu19}
Yu, Y., Smith, W.A.: Inverserendernet: Learning single image inverse rendering.
  In: Computer Vision and Pattern Recognition (CVPR) (2019)

\bibitem{KaiZhang2020}
Zhang, K., Riegler, G., Snavely, N., Koltun, V.: Nerf++: Analyzing and
  improving neural radiance fields. arXiv:2010.07492  (2020)

\bibitem{zhang2021nerfactor}
Zhang, X., Srinivasan, P.P., Deng, B., Debevec, P., Freeman, W.T., Barron,
  J.T.: Nerfactor: Neural factorization of shape and reflectance under an
  unknown illumination. ACM Trans. Graph.  (2021)

\end{thebibliography}
